\documentclass{eitnlp}
\eitlogo{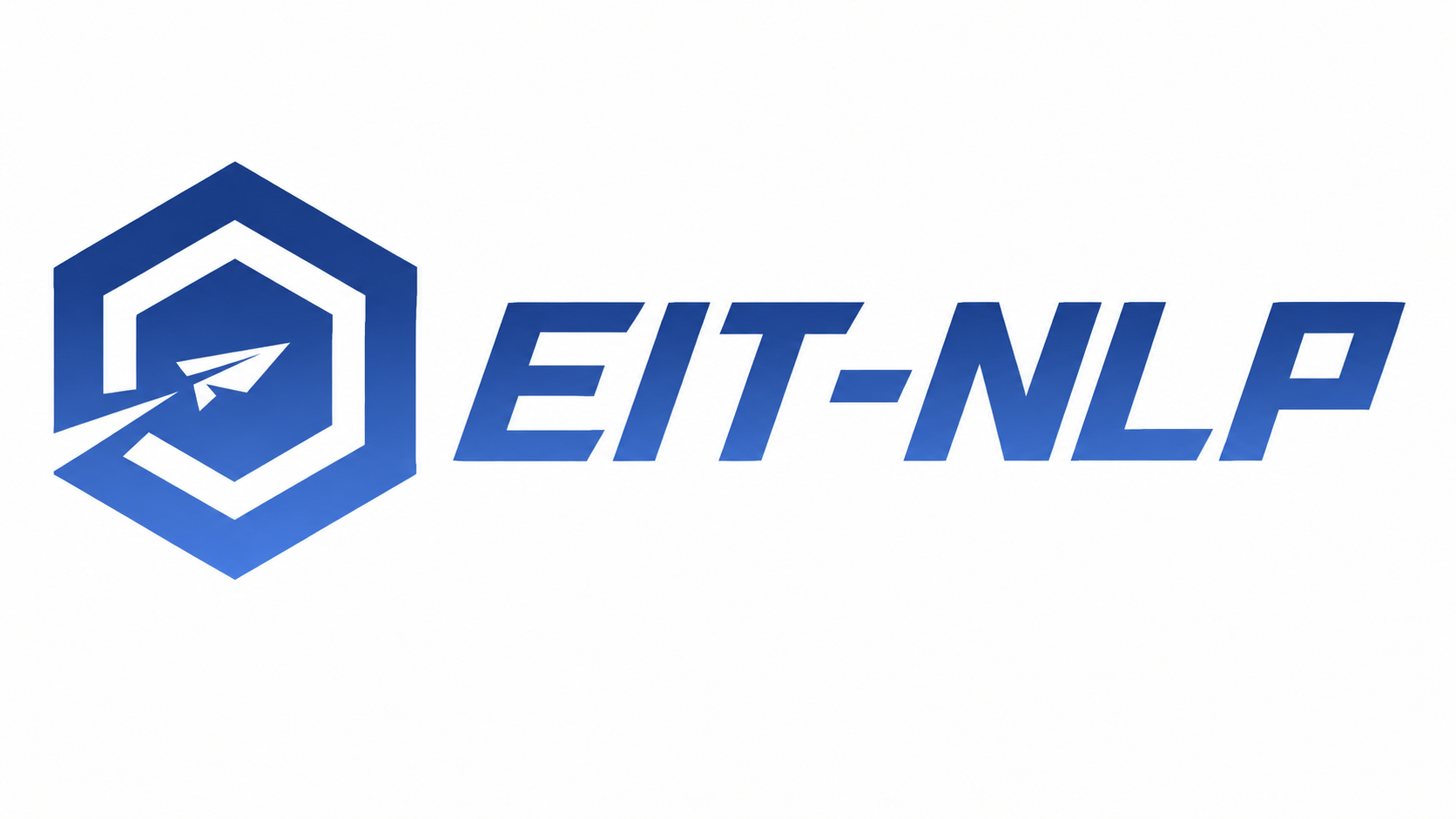}
\runningtitle{WIDE: Token-wise Structured Dynamic Width Pruning}

\usepackage[utf8]{inputenc} 
\usepackage[T1]{fontenc}    
\usepackage{hyperref}       
\usepackage{url}            
\usepackage{booktabs}       
\usepackage{amsfonts}       
\usepackage{nicefrac}       
\usepackage{microtype}      
\usepackage{xcolor}         

\usepackage{multirow}
\usepackage{amsmath}
\usepackage{amssymb}
\usepackage{mathtools}
\usepackage{amsthm}
\usepackage{wrapfig}
\usepackage{colortbl}
\usepackage{float}
\usepackage{placeins}
\usepackage{subcaption}
\usepackage{algorithm}
\usepackage{algpseudocode}
\usepackage{listings}
\usepackage{xcolor}

\lstdefinestyle{pytorch}{
    language=Python,
    basicstyle=\ttfamily\small,
    keywordstyle=\color{blue!70!black},
    commentstyle=\color{gray!80!black},
    stringstyle=\color{red!60!black},
    showstringspaces=false,
    breaklines=true,
    frame=single,
    rulecolor=\color{black!20},
    backgroundcolor=\color{gray!5},
    tabsize=4
}

\title{WIDE: Boosting Adaptive LLM Inference via Token-level Dynamic Width Pruning}
\newcommand{\med}{WIDE}

\author{
    Haozhe Hu\textsuperscript{1,2}\quad Hao Wu\textsuperscript{1,2}\quad
    Peiran Yin\textsuperscript{1,2}\quad Chao Han\textsuperscript{1}\\
    Yunpu Ma\textsuperscript{3}\quad Xiaoyu Shen\textsuperscript{1,\textdagger}
}
\affiliation{
    \textsuperscript{1}EIT-NLP Lab, Eastern Institute of Technology, Ningbo\\
    \textsuperscript{2}Shanghai Jiao Tong University\qquad
    \textsuperscript{3}Munich Center for Machine Learning, LMU Munich
}
\authornotes{\textsuperscript{\textdagger}Corresponding Author}
\contact{\href{mailto:email@domain}{Sophie17@sjtu.edu.cn; xyshen@eitech.edu.cn}}

\begin{document}

\maketitle
\begin{abstract}
Pruning is a promising approach for improving the efficiency of large language models (LLMs). Static structured pruning is hardware-friendly, but its input-agnostic computation allocation often degrades model quality at high sparsity. Dynamic pruning adapts computation to individual tokens, but coarse-grained layer-level routing limits allocation flexibility, and finer-grained decisions introduce irregular execution that can offset computational savings. To address these challenges, we present~\med, the first end-to-end differentiable framework that combines token-wise structured dynamic width pruning with GPU kernel co-design for both prefill and decoding. \med~uses differentiable routers to select GQA-aligned query-head groups and GPU-tile-aligned FFN-channel groups for each token. To execute these decisions efficiently, it reorders routing masks to cluster active tokens within each group, enabling hardware-agnostic block-level skipping and architecture-aware skipping of inactive loads and tensor-core operations. Experiments on the Llama3 and Qwen3 families demonstrate improved quality retention over representative static and dynamic pruning methods. At 50\% target routing sparsity, \med~improves Llama3.1-8B average zero-shot accuracy by up to 20.26 points over the strongest dynamic-depth baseline under calibration-only settings, while retaining 97.6\%/88.2\% of dense Qwen3-14B likelihood/generation performance after LoRA tuning. Separate latency benchmarks demonstrate end-to-end speedups of up to $1.68\times$ for prefill and $1.55\times$ for decoding over the dense baseline. These results demonstrate that pruning--kernel co-design pushes the frontier of token-wise dynamic structured pruning, combining improved quality retention with practical acceleration across both prefill and decoding. Our code is avaliable at \url{https://github.com/EIT-NLP/LLM-Pruning/tree/main/WIDE}.

\end{abstract}

\section{Introduction}
The rapid advancement of LLMs has enabled their widespread deployment across a broad range of real-world applications~\citep{KimiK25,DeepSeekV4,meng2026AgentHarness-Survey,ByteDance2026CUDAAgent,Nvidia2026AVO}. However, as models continue to scale, efficient LLM serving becomes increasingly challenging. Beyond system-level and hardware-specific optimizations~\citep{VLLM,SGLang,AWQ,FlashAttention4,tan2026vicostream}, model pruning has also evolved rapidly as a model-level approach to improving inference efficiency. By removing redundant structures along the depth and width dimensions, pruning reduces model size and computational cost, thereby offering a complementary path toward efficient LLM/VLM serving~\citep{cheng2024Pruning-Survey,frantar2023SparseGPT,ma2023LLMPruner,ashkboos2023SliceGPT,men2025ShortGPT,wu2025HiDrop,wu2026Survey, zheng2026geogr,wang2026mwop,liu2026vica}.

Currently, LLM pruning is dominated by static schemes, which remove predefined structures such as entire layers~\citep{kim2024ShortenedLLaMA}, attention or feed-forward network (FFN) sublayers~\citep{zhong2025BlockPruner}, rows or columns of weight matrices~\citep{li2025TyrthePruner}, and individual neurons~\citep{fang2024MaskLLM}. Once calibrated, these units are permanently removed, making static pruning simple to deploy and compatible with existing hardware backends. However, applying the same pruning decisions to every input often sacrifices model quality, especially under aggressive sparsity.

In contrast, dynamic pruning adapts computation to individual tokens through trainable routers~\citep{raposo2024MoD, zhao2025SkipGPT, han2025LFF} or activation thresholds~\citep{shrestha2025Polar, chen2025WINA}. Representative routing-based approaches make execution decisions at the layer or sublayer level, requiring each token to retain or skip an entire attention or FFN module. This coarse granularity limits their ability to preserve useful computation within a module while removing its redundant parts. Finer-grained width-level routing offers more flexible computation allocation, but introduces a systems challenge: different tokens may activate different subsets of heads or channels, producing irregular workloads that are difficult to execute efficiently on GPUs. The resulting synchronization and data-movement overheads can offset the savings from reduced computation~\citep{he2026SkipOPU, hu2026FLOPs}. Meanwhile, activation-based unstructured pruning primarily targets sparse matrix-vector operations during decoding, leaving efficient execution across both prefill and decoding insufficiently addressed. These limitations motivate a central question: \emph{How can token-wise dynamic pruning achieve finer-grained computation allocation while preserving the execution regularity needed for practical end-to-end acceleration?}

Motivated by these challenges, we present~\textbf{\med}. To the best of our knowledge, \med~is the first framework to combine token-wise dynamic width pruning with GPU kernel co-design for both prefill and decoding. Specifically, \med~groups attention heads and FFN channels according to GPU execution granularity and treats each group as the atomic pruning unit for each token, providing a trade-off between structured pruning granularity and execution flexibility. Furthermore, to translate these fine-grained decisions into acceleration, instead of explicitly materializing token-specific gathered tensors, \med~first performs mask reordering to cluster active tokens separately for each group into tile-aligned prefixes, \emph{converting irregular sparsity into CTA (Cooperative Thread Array)-level regularity.} Based on this representation, \med~kernels progressively eliminate unnecessary computation at multiple granularities: hardware-agnostic predicates remove fully inactive CTAs, while architecture-aware predicates further skip inactive activation loading and tensor-core computation. This design allows fine-grained dynamic sparsity to retain efficient GPU execution patterns and deliver practical acceleration for both prefill and decoding workloads.

Experiments on the Llama3 and Qwen3 families demonstrate improved quality retention over representative static and dynamic pruning methods. At 50\% target routing sparsity, \med~improves Llama3.1-8B average zero-shot accuracy by up to 20.26 points over the strongest dynamic-depth baseline under calibration-only settings, while retaining 97.6\%/88.2\% of dense Qwen3-14B likelihood/generation performance after LoRA tuning. Separate latency benchmarks show end-to-end speedups of up to 1.68$\times$ for prefill and 1.55$\times$ for decoding over the dense baseline. Kernel-level comparisons against naive gather--scatter implementations further show up to 200$\times$ speedup and a 1{,}000$\times$ reduction in peak memory usage. Our contributions are threefold:

\textbf{(1) We introduce~\med, the first end-to-end structured dynamic pruning framework for fine-grained token-wise width allocation.}
\med~formulates dynamic width pruning over GQA-aligned query-head groups and GPU-tile-aligned FFN-channel groups, with differentiable routers learning token-specific execution decisions. This extends structured routing beyond layer-level decisions to group-level allocation within attention and FFN modules.

\textbf{(2) We present the first unified framework that unlocks practical acceleration for fine-grained dynamic sparsity.}
To address synchronization and data-movement overheads, we combine mask reordering, block-level skipping, and architecture-aware intra-block predication, converting irregular token-wise selections into block-level regularity without materializing token-specific weight subsets. Dedicated fast paths support efficient decoding.

\textbf{(3) We advance the pruning frontier through pruning--kernel co-design.} Experiments demonstrate improved quality retention alongside practical acceleration across both prefill and decoding workloads. Further analysis characterizes how group size, recovery training, and GPU architecture shape the quality--efficiency trade-off, providing guidance for selecting pruning granularity.

\section{Related Works}
\paragraph{Static Pruning}
Among existing pruning methods, static pruning remains dominant. Given a small set of calibration samples, these methods estimate the importance of predefined units from Transformer layers to individual neurons, and permanently remove low-importance units until the target sparsity is reached. According to the pruning unit, existing work can be broadly categorized into depth and width pruning. \textbf{Depth} pruning removes computation along the layer dimension by calibration-based metric ranking, e.g., perplexity, Taylor~\citep{kim2024ShortenedLLaMA,zhong2025BlockPruner}, and surrogate-assisted Shapley values~\citep{ding2025CoopPruner}. \textbf{Width} pruning instead removes channels, neurons, or attention heads within layers. Methods include SliceGPT~\citep{ashkboos2023SliceGPT} and Týr-the-Pruner~\citep{li2025TyrthePruner} allocation row-wise sparsity under heuristic metrics, similar to depth pruning, while recent Deterministic Differentiable Pruning (DDP)~\citep{huang2026DDP} further learns differentiable structured masks under an $\ell_0$ sparsity constraint via augmented Lagrangian optimization. Beyond structural pruning, methods like Wanda~\citep{sun2023Simple} and HyperPrune~\citep{sun2025HyperPrune} further incorporate semi-structured or unstructured masks to prune individual weights. These static methods are deployment-friendly, but their input-agnostic pruning decisions limit their ability to allocate computation adaptively across tokens.

\paragraph{Dynamic Pruning}
Dynamic pruning allocates computation adaptively during inference instead of permanently removing a fixed set of parameters. Existing methods mainly focus on token-wise depth or sublayer routing with trainable decision modules, like Mixture-of-Depths (MoD)~\citep{raposo2024MoD}, D-LLM~\citep{jiang2024DLLM}, SkipGPT~\citep{zhao2025SkipGPT}, and LFF~\citep{han2025LFF}. PolarSparse and FastForward instead introduce the dynamic width pruning for batch decode / prefill only~\citep{gautam2026Fast,shrestha2025Polar}. As for router-free decision, methods such as WINA~\citep{chen2025WINA} incorporate dynamic unstructured pruning with activation-based threshold for decoding acceleration. While these approaches improve adaptive computation, they mainly operate at coarse granularity or specific scenario. 
\med~instead formulates token-wise structured width pruning over query-head groups and FFN-channel groups, with differentiable routers learning group decisions end to end. It couples this fine-grained allocation with GPU-oriented group-wise skipping for both prefill and decoding.
\begin{figure}[!t]
    \centering
    \includegraphics[width=1.0\linewidth, trim=10 300 10 300, clip]{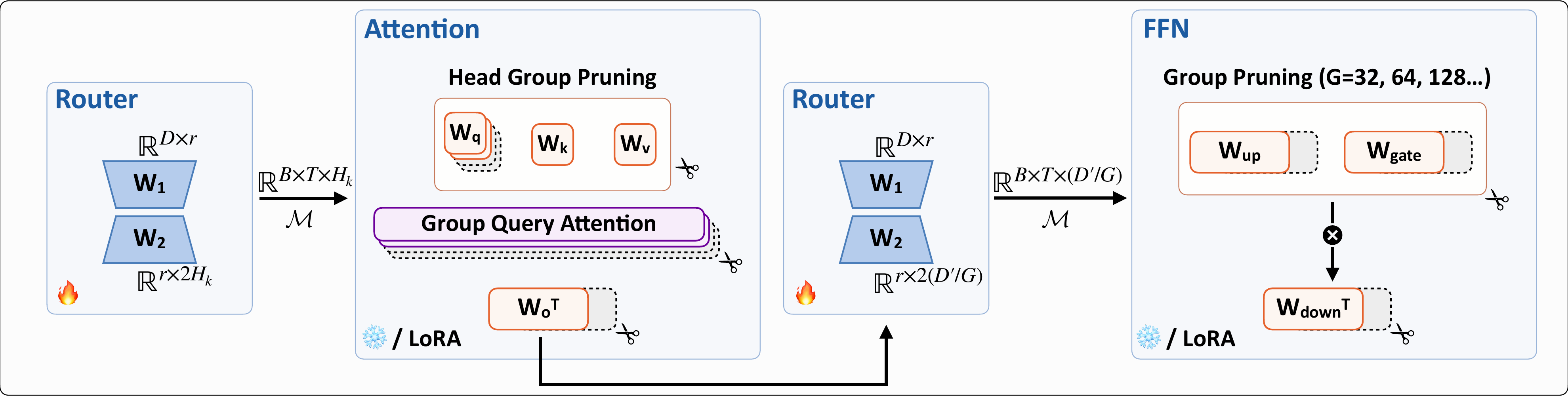}
    \caption{The pruning dimension of \med~in the attention and FFN layer, with $\mathbf{W}_q$, $\mathbf{W}_{\text{up}}$, and $\mathbf{W}_{\text{gate}}$ pruned over out features (N-axis), and $\mathbf{W}_o$, $\mathbf{W}_{\text{down}}$ over in features (K-axis). Each router produces \(2N_G\) logits for \(N_G\) binary decisions to align with the Gumbel Softmax.}
    \label{fig:overview}
\end{figure}

\section{\med: Preliminary, Training, and Inference Design}
As previously introduced, \med~is a dynamic width pruning framework that selectively activates head / channel groups under token-wise routing. In this section, we first define the minimal atomic pruning unit of \med, then describe the designs for training and GPU-side execution.

\subsection{Preliminaries}
Consider a standard Transformer layer with input token embeddings $\textbf{X} \in \mathbb{R}^{B \times T \times D}$ with $B$ for batch size, $T$ for token length, and $D$ for hidden size, \med~follows the granularity of SkipGPT~\citep{zhao2025SkipGPT}, which decomposes the entire layer into two computation parts: attention and FFN. In attention, we use $T_q$, $T_k$ to denote the token length of query and key value, $H_q$, $H_k$ denote the number of query / key value attention heads, and $d = \frac{D}{H_q}$ for the head size. For FFN, $D^\prime$ denotes the intermediate size of FFN's up scaling. The overall pruning granularity of \med~is shown in Figure~\ref{fig:overview}, where each token allocates its computation budget based on the atomic pruning group size $G$.

\paragraph{Attention Pruning Unit}
For the attention layer, considering the compatibility with the standard FlashAttention tiling scheme that launches over $B$, $H_q$, $T_q$ dimensions and consumes one head at each step~\citep{FlashAttention2}, we assume the head as the minimal pruning atom to skip $d$ or multiple-$d$ under each token routing. Such pruning strategy can be equivalently represented with a mask of the form:
\begingroup
\small
\setlength{\abovedisplayskip}{3pt}
\setlength{\belowdisplayskip}{3pt}
\setlength{\abovedisplayshortskip}{3pt}
\setlength{\belowdisplayshortskip}{3pt}
\begin{align}
    & [\textbf{Q}, \textbf{K}, \textbf{V}] = [\textbf{X}\mathbf{W}_q^\top, \textbf{X}_k\mathbf{W}_k^\top, \textbf{X}_v\mathbf{W}_v^\top]\\
    & \textbf{X}^\prime = \mathsf{S}\left(\underbrace{\text{Attn.}(\textbf{Q}, \textbf{K}, \textbf{V})}_{\mathbb{R}^{B\times T_q\times H_q\times d}}\odot_G \mathcal{M}_{\text{attn}}\right)\mathbf{W}_o^\top\quad \text{where}\ \mathcal{M}_{\text{attn}}\in \{0,1\}^{B\times T_q\times H_k},\ G = \frac{H_q}{H_k}\times d
\end{align}
\endgroup
Here $\odot_G$ denotes group-wise gating, $\mathsf{S}$ denotes the reshape operation. The outer mask $\mathcal{M}_{\text{attn}}$ also leaves $\mathbf{W}_k$ and $\mathbf{W}_v$ unpruned to preserve the dense KV cache. Additionally, since recent LLMs predominantly adopt GQA~\citep{GQA}, \med~sets attention's $G_{\text{attn}}$ to the head group size $\frac{H_q}{H_k}\times d$ for preserving compatibility with existing GQA decoding optimizations (e.g., XQA~\citep{TensorRTLLM,FlashInfer}) while minimizing the degradation of pruning flexibility.

\paragraph{FFN Pruning Unit}
Unlike attention layers, FFN layers comprise only general matrix multiplication (GEMM) and element-wise operations. For a standard GEMM operation with its left-hand-side operand (LHS) $\textbf{A}\in \mathbb{R}^{M\times K}$ and right-hand-side operand (RHS) $\textbf{B}\in \mathbb{R}^{N\times K}$, the tiling strategy typically slices all three logical dimensions $M$, $N$, and $K$, with tile sizes $(BM, BN, BK) \in \mathcal{T}^3,\quad \mathcal{T} = \{2^k \mid k \in \mathbb{Z}_{\geqslant 0}\}$. Based on this, \med~adopts commonly used tiling sizes as the $G$ for FFN to prune the intermediate dimension $D^\prime$ into contiguous groups. This dimension corresponds to the out-feature in $\mathbf{W}_{\text{up}}$/$\mathbf{W}_{\text{gate}}$ and the in-feature in $\mathbf{W}_{\text{down}}$, ensuring a single mask matrix to represent the entire FFN layer pruning:
\begingroup
\small
\setlength{\abovedisplayskip}{3pt}
\setlength{\belowdisplayskip}{3pt}
\setlength{\abovedisplayshortskip}{3pt}
\setlength{\belowdisplayshortskip}{3pt}
\begin{equation}
    \textbf{X}^\prime = \left(\underbrace{(\phi(\textbf{X}\mathbf{W}_{\text{gate}}^\top)\odot (\textbf{X}\mathbf{W}_{\text{up}}^\top))}_{\mathbb{R}^{B\times T\times (D^\prime / G, G)}}\odot_G \mathcal{M}_{\text{ffn}}\right)\mathbf{W}_{\text{down}}^\top\quad \text{where}\ \mathcal{M}_{\text{ffn}}\in \{0, 1\}^{B\times T\times \frac{D^\prime}{G}},\ G \in \mathcal{T}
\end{equation}
\endgroup
where $\phi$ denotes element-wise activation function. The FFN unit offers more flexibility in choosing $G$ than the attention unit, as its available group sizes are drawn from the tiling-size set $\mathcal{T}$, trading off pruning quality ($G\downarrow$) against potential acceleration ($G\uparrow$). Therefore, unless otherwise specified, $G$ refers to the FFN group size $G_{\text{ffn}}$ hereafter.

\paragraph{Routing Components}
Given the pruning units and masks above, \med~uses a bottleneck router to map each input activation to $N_G$ binary decisions, supporting varying dimensions and Gumbel Softmax sampling during training:
\begingroup
\small
\setlength{\abovedisplayskip}{3pt}
\setlength{\belowdisplayskip}{3pt}
\setlength{\abovedisplayshortskip}{3pt}
\setlength{\belowdisplayshortskip}{3pt}
\begin{align}
    &\mathbf{R} = \mathsf{S}\left(\textbf{X}\textbf{W}_1 \textbf{W}_2\right)\in \mathbb{R}^{B\times T\times N_G\times 2}\quad \text{where}\ \textbf{W}_1\in \mathbb{R}^{D\times r},\ \textbf{W}_2\in \mathbb{R}^{r\times 2N_G}\\
    &\mathcal{M}=
    \begin{cases}
    \mathbf{1}\!\left[
    \text{argmax}_{c\in\{0,1\}}\mathbf{R}_{\ldots,c}=0\right],
    & \text{when inference},\\
    \operatorname{GumbelSoftmax}_{\tau}^{\text{hard}}
    (\mathbf{R})_{\ldots,0},
    & \text{when training with straight-through estimator (STE)},
    \end{cases}
\end{align}
\endgroup
where $\tau$ is the temperature, $r\ll D$ is the bottleneck rank, and $N_G$ is $D/G$ for attention and $D^\prime/G$ for FFN, with class $0$ denotes execution and class $1$ for skipping.

\paragraph{Two-stage Training}
To fully unlock the potential of \med, a two-stage training pipeline is introduced to restore model's original performance. In the first stage, called \textbf{Router training} (S1), all parameters except the router remain frozen, with calibration data employed only to optimize the routing components. In the next optional \textbf{LoRA tuning} (S2) stage, LoRA adapters are added into base models' linear modules to further support performance recovery, with routers remain trainable. Both stages utilize not only the standard language modeling loss but also an auxiliary sparsity loss to ensure the pruning meets the target sparsity. For a model with $L$ layers, the overall objective is:
\begingroup
\small
\setlength{\abovedisplayskip}{3pt}
\setlength{\belowdisplayskip}{3pt}
\setlength{\abovedisplayshortskip}{3pt}
\setlength{\belowdisplayshortskip}{3pt}
\begin{equation}
    \mathcal{L}
    =
    \mathcal{L}_{\text{LM}}
    +\alpha\left\lvert
    SP-\frac{1}{2L}\sum_{i=1}^{L}
    \left(
    \left(1-\frac{1}{\lvert\mathcal{M}_{\text{attn}}^i\rvert}\sum\mathcal{M}_{\text{attn}}^i\right)
    +
    \left(1-\frac{1}{\lvert\mathcal{M}_{\text{ffn}}^i\rvert}\sum\mathcal{M}_{\text{ffn}}^i\right)
    \right)
    \right\rvert
    \label{eq:loss}
\end{equation}
\endgroup
where $\alpha$ is a hyperparameter controlling the weight of the sparsity loss, $\lvert\mathcal{M}\rvert$ denotes the total element number of the mask. Here, $SP$ denotes the target routing sparsity. Since the total computational proportion of attention varies with $T$, we follow SkipGPT and weight attention and FFN equally when computing the overall sparsity.
\begin{figure}[!t]
    \centering
    \includegraphics[width=1.0\linewidth, trim=10 70 55 10, clip]{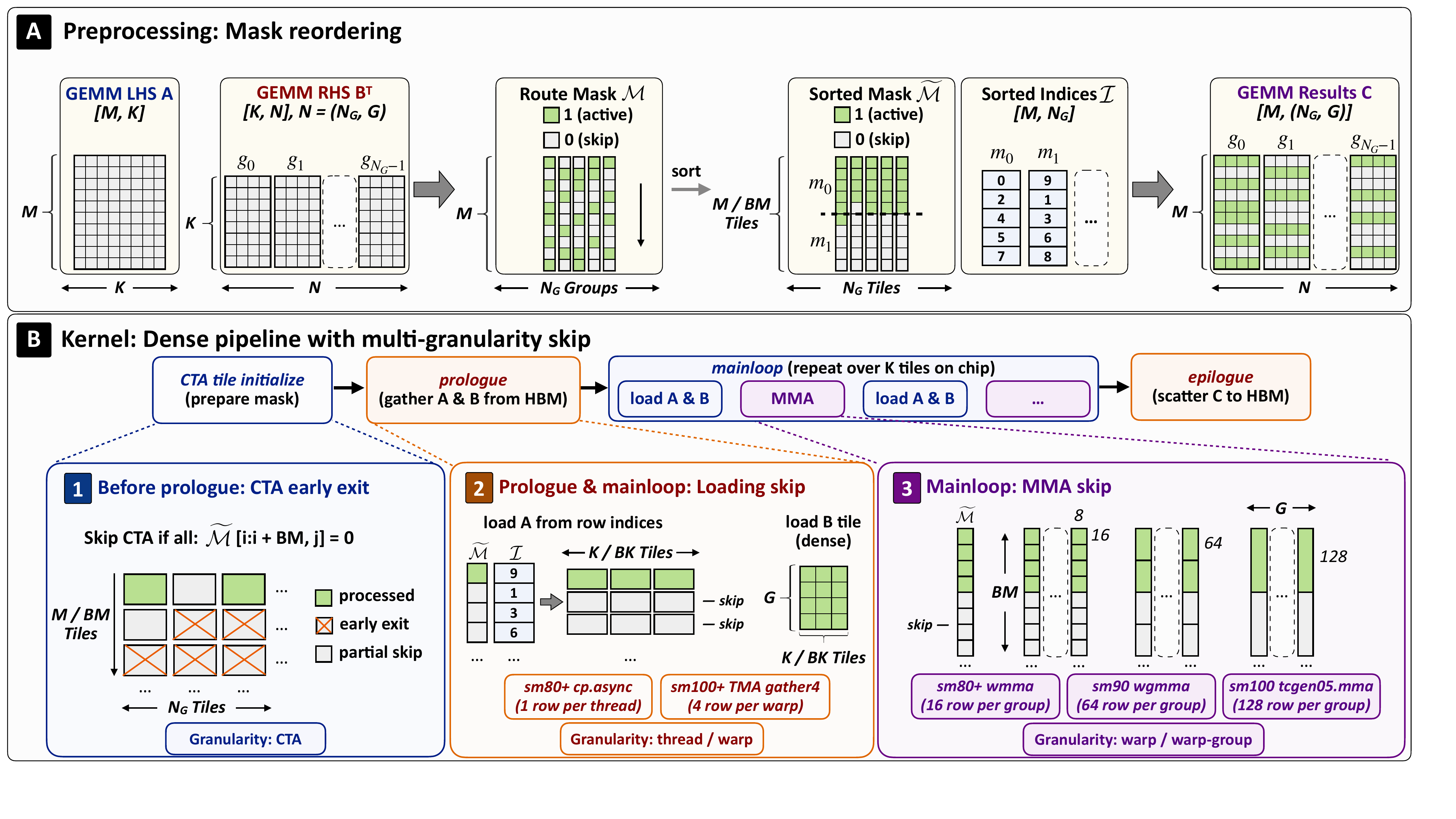}
    \caption{Workflow of \med~for GEMM N-axis pruning (without split-K). \textbf{(a)} Given group size $G$, \med~generates a mask matrix of shape $[M, N_G]$ and sorts it along the $M$ dimension, so that active rows of each group are clustered within $BM$-token tiles. \textbf{(b)} Inside the kernel, \med~first skips fully inactive $[BM, G]$ blocks. For blocks with active entries, it follows the standard GEMM pipeline and applies architecture-dependent intra-block skipping during \textbf{A} loading and MMA execution.}
    \label{fig:kernel}
\end{figure}

\subsection{Inference-time Group Skipping with Unified Mask Reordering}
Unlike MoE-style routing, dynamic pruning does not assume a fixed compute budget for each token. A naive implementation therefore typically needs to explicitly convert routing masks into exact active indices and then use gather-scatter operations to feed only active parts into kernels. Let $\mathbf{C}=\mathbf{A}\mathbf{B}^{\top}$, where $\mathbf{A}\in\mathbb{R}^{M\times K}$, $\mathbf{B}\in\mathbb{R}^{N\times K}$, and $\mathbf{C}\in\mathbb{R}^{M\times N}$. This strategy introduces two major overheads:

\textbf{(1). Device-to-host synchronization.}
The active token layout must be derived from input-dependent masks before constructing the compact gathered matrix, which can trigger device-to-host synchronization and break CUDA Graph execution.\newline
\textbf{(2). Extra memory movement and storage.}
For finer-grained dynamic width pruning, each token may select a different subset of $\mathbf{B}$. A naive N-axis implementation therefore incurs traffic of $MK+3pMNK+3pMN+MN$, while its K-axis counterpart incurs $3pMK+3pMNK+MN$, both require intermediate storage of $\mathcal{O}(pMNK)$ for token-specific weight subsets (see Appendix~\ref{apdx:memory-accounting}).

Under this notation, \med~uses a unified mask-reordering step with fused gather-scatter kernels while retaining the original tile-based pipeline, as illustrated in Fig.~\ref{fig:kernel}. Specifically, dynamic routing is represented by a binary mask $\mathcal{M}\in\{0,1\}^{M\times N_G}$, where $N_G=1$ for M-axis routing (i.e., dynamic depth) and $N_G=D_{\text{p}}/G$ for width pruning over the pruned feature dimension $D_{\text{p}}$. For each group $g$, sorting the mask along $M$ produces a reordered mask $\widetilde{\mathcal{M}}_{:,g}$ and physical row indices $\mathcal{I}_{:,g}$:
\begingroup
\small
\setlength{\abovedisplayskip}{3pt}
\setlength{\belowdisplayskip}{3pt}
\setlength{\abovedisplayshortskip}{3pt}
\setlength{\belowdisplayshortskip}{3pt}
\begin{align}
    (\widetilde{\mathcal{M}}_{:,g}, \mathcal{I}_{:,g})
    =
    \operatorname{SortDesc}(\mathcal{M}_{:,g})
\end{align}
\endgroup
This clusters and packs active rows into a contiguous prefix, while $\mathcal{I}_{:,g}$ gathers activation rows and scatters outputs back to their original positions. After sorting, all but the boundary CTA of each group become fully active or inactive. Active rows within each admitted CTA therefore share a group-aligned weight tile $\mathbf{B}_{\mathrm{tile}}\in\mathbb{R}^{BN\times BK}$, enabling multi-granularity predicated skipping in the dense-style pipeline without token-specific weight materialization. \med~then applies unified skipping predicates over progressively finer row groups. Let
\begin{align}
    S_0 = BM, \qquad S_1 = S_{\text{ld}}, \qquad S_2 = S_{\text{mma}},
\end{align}
where $S_0$ is the CTA-level block size, while $S_{\text{ld}}$ and $S_{\text{mma}}$ are architecture-dependent granularities for $\mathbf{A}$ loading and MMA execution. For level $\ell\in\{0,1,2\}$, group $g$, and row block $m$, \med~computes
\begingroup
\small
\setlength{\abovedisplayskip}{3pt}
\setlength{\belowdisplayskip}{3pt}
\setlength{\abovedisplayshortskip}{3pt}
\setlength{\belowdisplayshortskip}{3pt}
\begin{align}
    s^{(\ell)}_{m,g}
    =
    \bigvee_{i=mS_\ell}^{\min((m+1)S_\ell,M)-1}
    \widetilde{\mathcal{M}}_{i,g}
\end{align}
\endgroup
If $s^{(\ell)}_{m,g}=0$, the corresponding unit at level $\ell$ is skipped; otherwise it is executed. This single predicate instantiates three stages: $\ell=0$ performs hardware-agnostic CTA early exit, $\ell=1$ skips global memory loading for inactive rows in $\mathbf{A}$, and $\ell=2$ skips inactive MMA fragments. Under this abstraction, the proposed \textbf{GEMM-MN}, \textbf{GEMM-K}, and \textbf{Attention} kernels cover $\mathbf{W}_q$/$\mathbf{W_{\text{up}}}$/$\mathbf{W_{\text{gate}}}$ projections, $\mathbf{W}_o$/$\mathbf{W_{\text{down}}}$ projections, and FlashAttention operations with customized routing group, allowing \med~to exploit the acceleration potential brought by sparsity while preserving its dense-style mainloops. At the kernel level, GEMM-K naturally requires K-axis accumulation to be partitioned according to $G$. We place multiple K-axis routing groups within a CTA and process them through an outer loop to balance atomic-add pressure. For decoding, reordering-free GEMV/FlashDecoding fast paths have been introduced, which directly perform CTA-level skipping from the original mask. The detailed implementation for each kernel is available at Appendix~\ref{apdx:cute-kernel-details}.

\section{Experiments}
\subsection{Settings}
\paragraph{Training}
We use Llama3.1-8B and Llama3.2-3B~\citep{Llama3} as the main backbone models, along with RedPajama-1T~\citep{redpajama} as the corpus for calibration-only (S1) and LoRA tuning (S2) stages. The weight $\alpha$ in Equation~\ref{eq:loss} is set to 20, following SkipGPT. As for the Qwen3 family~\citep{Qwen3}, Infinity-Instruct~\citep{InfinityInstruct} is introduced as the calibration set to retain the capability of instruction following. More details are available at Appendix~\ref{sec:Implementation Details for Training and Evaluation}.
\begin{table}[!t]
    \centering
    \caption{Performance on \textcolor{gray}{\textbf{static depth}}, \textcolor{orange!50}{\textbf{static width}}, and \textcolor{cyan!50}{\textbf{dynamic}} pruning methods under Llama3 and various target sparsity, training stages, and tasks. The S1 denotes calibration-only / router training stage, and S2 denotes LoRA tuning stage. The \textbf{WT2} denotes WikiText2 perplexity ($\downarrow$), \textbf{Avg. Acc.} reports average accuracy ($\uparrow$) on 7 likelihood tasks. $G$ in \med~($G$) denotes the FFN group size.}
    \aboverulesep=0ex
    \belowrulesep=0ex
    \renewcommand{\arraystretch}{1.2}
    \setlength{\tabcolsep}{2pt}
    \resizebox{1.0\linewidth}{!}{
        \begin{tabular}{l|cc|cc|cc|cc|cc|cc}
            \toprule
            \multirow{4}{*}{Methods} & \multicolumn{8}{c|}{Llama3.1-8B} & \multicolumn{4}{c}{Llama3.2-3B}\\
            \cline{2-9}\cline{10-13}
            & \multicolumn{4}{c|}{25\% sparsity} & \multicolumn{4}{c|}{50\% sparsity} & \multicolumn{4}{c}{50\% sparsity}\\
            \cline{2-5}\cline{6-9}\cline{10-13}
            & \multicolumn{2}{c|}{S1} & \multicolumn{2}{c|}{S2} & \multicolumn{2}{c|}{S1} & \multicolumn{2}{c|}{S2} & \multicolumn{2}{c|}{S1} & \multicolumn{2}{c}{S2}\\
            \cline{2-3}\cline{4-5}\cline{6-7}\cline{8-9}\cline{10-11}\cline{12-13}
            & WT2 & Avg. Acc. & WT2 & Avg. Acc. & WT2 & Avg. Acc. & WT2 & Avg. Acc. & WT2 & Avg. Acc. & WT2 & Avg. Acc.\\
            \midrule
            Dense & 7.71 & 71.55 (100.00\%) & 7.91 & 71.36 (99.73\%) & 7.71 & 71.55 (100.00\%) & 7.91 & 71.36 (99.73\%) & 9.77 & 64.91 (100.00\%) & 9.86 & 65.00 (100.14\%) \\
            \midrule
            \rowcolor{gray!15}
            Shortened-ppl & 25.84 & 47.65 (66.59\%) & 12.33 & 57.25 (80.01\%) & 473.15 & 39.62 (55.37\%) & 22.47 & 47.26 (66.05\%) & 676.05 & 38.23 (58.89\%) & 29.75 & 45.83 (70.61\%) \\
            \rowcolor{gray!15}
            Shortened-taylor & 30.07 & 49.30 (68.90\%) & 12.12 & 59.83 (83.63\%) & 4.86e+7 & 35.99 (50.30\%) & 22.30 & 49.97 (69.83\%) & 2.32e+5 & 39.35 (60.63\%) & 31.50 & 45.84 (70.61\%) \\
            \rowcolor{gray!15}
            CoopPruner & 22.96 & 53.33 (74.54\%) & 12.03 & 61.25 (85.61\%) & 503.86 & 39.28 (54.90\%) & 22.66 & 49.63 (69.36\%) & 1.79e+5 & 36.62 (56.41\%) & 42.69 & 41.82 (64.43\%) \\
            \rowcolor{orange!15}
            SliceGPT & 22.57 & 53.37 (74.59\%) & 14.88 & 60.72 (84.87\%) & 85.81 & 38.71 (54.10\%) & 28.76 & 46.01 (64.31\%) & 96.69 & 37.62 (57.96\%) & 31.13 & 42.76 (65.87\%) \\
            \rowcolor{orange!15}
            T\'yr-the-Pruner & 12.89 & 61.43 (85.86\%) & 11.49 & 64.20 (89.73\%) & 59.88 & 45.47 (63.55\%) & 37.99 & 53.46 (74.72\%) & 87.37 & 42.60 (65.63\%) & 53.11 & 45.69 (70.38\%) \\
            \rowcolor{orange!15}
            DDP & 12.41 & 64.35 (89.93\%) & 12.10 & 64.24 (89.79\%) & 21.83 & 53.05 (74.14\%) & 19.52 & 54.64 (76.36\%) & 31.20 & 48.03 (74.00\%) & 25.66 & 48.88 (75.30\%) \\
            \rowcolor{cyan!15}
            D-LLM & 69.45 & 38.89 (54.35\%) & 10.29 & 66.84 (93.42\%) & 504.48 & 36.32 (50.77\%) & 19.31 & 56.07 (78.36\%) & 3877.53 & 34.99 (53.91\%) & 193.97 & 38.66 (59.56\%) \\
            \rowcolor{cyan!15}
            SkipGPT & 15.63 & 51.68 (72.23\%) & \underline{9.87} & 69.48 (97.11\%) & 96.04 & 42.52 (59.42\%) & 13.90 & 61.60 (86.10\%) & 156.58 & 36.74 (56.60\%) & 18.91 & 54.33 (83.70\%) \\
            \rowcolor{cyan!15}
            \med~(32) & \textbf{8.49} & \textbf{70.06 (97.92\%)} & \textbf{8.61} & \textbf{70.19 (98.09\%)} & \underline{14.15} & \textbf{61.84 (86.43\%)} & \underline{11.99} & \textbf{64.82 (90.60\%)} & \textbf{17.28} & \textbf{57.00 (87.81\%)} & 15.28 & \textbf{58.77 (90.53\%)} \\
            \rowcolor{cyan!15}
            \med~(64) & 10.37 & \underline{64.89 (90.69\%)} & 10.32 & \underline{69.83 (97.59\%)} & 14.96 & 61.18 (85.50\%) & 12.05 & \underline{64.65 (90.36\%)} & 19.02 & \underline{56.40 (86.89\%)} & \textbf{15.17} & \underline{58.36 (89.90\%)} \\
            \rowcolor{cyan!15}
            \med~(128) & \underline{10.04} & 63.65 (88.96\%) & 12.51 & 69.53 (97.18\%) & \textbf{12.57} & \underline{61.49 (85.94\%)} & \textbf{11.43} & 64.34 (89.92\%) & \underline{18.11} & 55.42 (85.38\%) & \underline{15.26} & 58.11 (89.52\%) \\
            \bottomrule
        \end{tabular}
    }
    \label{tbl:performance_llama}
\end{table}
\begin{table}[!t]
    \centering
    \caption{Performance on Qwen3-14B with various target sparsity. \textbf{Like. Acc.} denotes average accuracy in 8 likelihood tasks, and \textbf{Gen. Acc.} denotes average accuracy in 3 generation tasks.}
    \aboverulesep=0ex
    \belowrulesep=0ex
    \renewcommand{\arraystretch}{1.2}
    \setlength{\tabcolsep}{2pt}
    \resizebox{1.0\linewidth}{!}{
        \begin{tabular}{l|cc|cc|cc|cc}
            \toprule
            \multirow{4}{*}{Methods} & \multicolumn{8}{c}{Qwen3-14B}\\
            \cline{2-5}\cline{6-9}
            & \multicolumn{4}{c|}{25\% sparsity} & \multicolumn{4}{c}{50\% sparsity} \\
            \cline{2-3}\cline{4-5}\cline{6-7}\cline{8-9}
            & \multicolumn{2}{c|}{S1} & \multicolumn{2}{c|}{S2} & \multicolumn{2}{c|}{S1} & \multicolumn{2}{c}{S2} \\
            \cline{2-3}\cline{4-5}\cline{6-7}\cline{8-9}
            & Like. Acc. & Gen. Acc. & Like. Acc. & Gen. Acc. & Like. Acc. & Gen. Acc. & Like. Acc. & Gen. Acc. \\
            \midrule
            Dense & 73.57 (100.00\%) & 84.79 (100.00\%) & 76.16 (103.52\%) & 83.86 (98.91\%) & 73.57 (100.00\%) & 84.79 (100.00\%) & 76.16 (103.52\%) & 83.86 (98.91\%) \\
            \midrule
            \rowcolor{gray!15}
            Shortened-ppl & 44.55 (60.55\%) & 4.33 (5.11\%) & 59.90 (81.42\%) & 40.55 (47.82\%) & 34.39 (46.74\%) & 3.22 (3.79\%) & 44.91 (61.05\%) & 5.68 (6.69\%) \\
            \rowcolor{gray!15}
            Shortened-taylor & 36.76 (49.97\%) & 4.93 (5.81\%) & 65.26 (88.70\%) & 67.09 (79.13\%) & 35.50 (48.26\%) & 3.79 (4.47\%) & 46.62 (63.37\%) & 20.03 (23.63\%) \\
            \rowcolor{gray!15}
            CoopPruner & 44.23 (60.12\%) & 5.92 (6.98\%) & 62.89 (85.48\%) & 52.20 (61.56\%) & 35.66 (48.47\%) & 4.45 (5.25\%) & 47.08 (63.99\%) & 11.44 (13.50\%) \\
            \rowcolor{orange!15}
            DDP & 65.26 (88.71\%) & 73.23 (86.37\%) & 68.77 (93.48\%) & 74.62 (88.00\%) & 53.00 (72.04\%) & 52.20 (61.56\%) & 55.74 (75.77\%) & 54.88 (64.72\%) \\
            \rowcolor{cyan!15}
            SkipGPT & 61.63 (83.77\%) & 38.34 (45.21\%) & 71.96 (97.80\%) & 76.67 (90.42\%) & 35.15 (47.77\%) & 8.09 (9.54\%) & 57.20 (77.74\%) & 68.12 (80.34\%) \\
            \rowcolor{cyan!15}
            \med~(32) & 72.70 (98.82\%) & \underline{77.98 (91.96\%)} & 74.01 (100.59\%) & \underline{80.89 (95.40\%)} & 68.95 (93.71\%) & \textbf{71.48 (84.31\%)} & 71.26 (96.86\%) & 73.11 (86.22\%) \\
            \rowcolor{cyan!15}
            \med~(64) & \textbf{73.19 (99.49\%)} & \underline{77.98 (91.97\%)} & \underline{74.22 (100.88\%)} & 80.80 (95.29\%) & \textbf{70.15 (95.35\%)} & 69.82 (82.35\%) & \underline{71.34 (96.96\%)} & \textbf{75.26 (88.76\%)} \\
            \rowcolor{cyan!15}
            \med~(128) & \underline{73.11 (99.37\%)} & \textbf{78.57 (92.67\%)} & \textbf{74.33 (101.03\%)} & \textbf{81.46 (96.08\%)} & \underline{69.24 (94.11\%)} & \underline{70.81 (83.52\%)} & \textbf{71.82 (97.62\%)} & \underline{74.85 (88.28\%)} \\
            \bottomrule
        \end{tabular}
    }
    \label{tbl:performance_qwen}
\end{table}

\paragraph{Evaluation}
For evaluating the capability of pruned models, the following metrics are included: perplexity on WikiText2~\citep{wikitext2}, and zero-shot accuracy on ARC-Easy, ARC-Challenge~\citep{arc}, BoolQ~\citep{boolq}, WinoGrande~\citep{winogrande}, PIQA~\citep{piqa}, OpenBookQA~\citep{openbookqa}, and HellaSwag~\citep{hellaswag}. For Qwen3 family, 5-shot MMLU~\citep{mmlu} is added as an extended likelihood task, with additional IFEval~\citep{ifeval}, MBPP+~\citep{mbpp}, and GSM8K~\citep{gsm8k} for evaluating the generation capability. The average accuracy and performance retention ratio are computed over the selected likelihood / generation tasks based on the dense model.

\paragraph{Inference Implementation}
For kernel-level throughput evaluation, we use the Triton benchmark interface~\citep{Triton} and report TFLOPs under CUDA Graph replay and random mask. For end-to-end acceleration, an ELANA-style profiling tool~\citep{chiang2025ELANA} is introduced to measure Time-To-First-Token (TTFT) and Time-Per-Output-Token (TPOT) under CUDA Graph execution. We implement \med~kernels in the tile-level Domain Specific Languages (DSLs), e.g., Triton~\citep{Triton} and TileLang~\citep{Tilelang}, as well as the lower-level CuTe template library in CUTLASS. All inference experiments are mainly conducted on GPUs with sm120 (NVIDIA RTX 5090) and sm80 (A100-SMX4-40G) architectures.

\paragraph{Baselines}
We compare \med~with representative pruning methods, including static depth pruning (Shortened LLaMA~\citep{kim2024ShortenedLLaMA}, CoopPruner~\citep{ding2025CoopPruner}), static width pruning (SliceGPT~\citep{ashkboos2023SliceGPT}, T\'yr-the-Pruner~\citep{li2025TyrthePruner}, DDP~\citep{huang2026DDP}), and dynamic pruning (D-LLM~\citep{jiang2024DLLM}, SkipGPT~\citep{zhao2025SkipGPT}). We also compare against Dense-to-MoE~\citep{pei2026CMoE}, semi-structured pruning~\citep{sun2025HyperPrune}, and activation-based dynamic pruning~\citep{shrestha2025Polar,chen2025WINA} in Appendix~\ref{sec:Additional Experimental Results}. Baseline training details are provided in Appendix~\ref{sec:Implementation Details for Training and Evaluation}.

\subsection{Main Results}
\paragraph{Benchmarks on Llama3-family and Qwen3}
Table~\ref{tbl:performance_llama} and~\ref{tbl:performance_qwen} compare pruning methods on the Llama3 / Qwen3 families at 25\% and 50\% sparsity. \med~achieves the strongest performance across both budgets, with the largest gains under 50\% sparsity and calibration-only settings. On Llama3, \med~outperforms the strongest static width and dynamic depth baselines by up to 18.68\% and 55.14\%, respectively. After LoRA tuning, the gains remain 20.23\% and 8.17\%. Broader comparisons with Dense-to-MoE, semi-structured pruning, and activation-based unstructured pruning (Table~\ref{tbl:additional_baseline_bench}) further show that \med~remains competitive against methods with even finer pruning granularity. On Qwen3-14B, \med~further retains nearly lossless likelihood-task performance after tuning, together with 96\% and 89\% of dense generation-task performance at 25\% and 50\% sparsity, respectively.

\begin{wraptable}{r}{0.35\linewidth}
    \centering
    \vspace{-15pt}
    \caption{Pruned models' final-layer error and similarity to dense Qwen3-14B at 50\% sparsity. \med~is closer in both stages.}
    \aboverulesep=0ex
    \belowrulesep=0ex
    \renewcommand{\arraystretch}{1.2}
    \setlength{\tabcolsep}{3pt}
    \resizebox{1.0\linewidth}{!}{
        \begin{tabular}{l|cc}
            \toprule
            Methods & Relative $\ell_2\downarrow$ & Cos. Sim.$\uparrow$ \\
            \midrule
            \multicolumn{3}{c}{S1} \\
            \midrule
            SkipGPT & 1.3090 & 0.7337 \\
            \med~(128) & \textbf{0.5778} & \textbf{0.8556} \\
            \midrule
            \multicolumn{3}{c}{S2} \\
            \midrule
            SkipGPT & 0.6642 & 0.8447 \\
            \med~(128) & \textbf{0.5510} & \textbf{0.8506} \\
            \bottomrule
        \end{tabular}
    }
    \vspace{-15pt}
    \label{tbl:layer_wise_error}
\end{wraptable}
\paragraph{Benefits from Finer-grained Routing}
Beyond the end-to-end task-level advantage, \med~also benefits from its finer-grained decisions in reducing layer-wise error accumulation. As shown in Table~\ref{tbl:layer_wise_error} and Appendix~\ref{sec:Additional Experimental Results}, with the dense model as the reference, \med~consistently yields lower errors than SkipGPT both at individual layers and at the final-layer output, which also helps explain its stronger performance over SkipGPT.
\begin{figure}[!t]
    \centering
    \includegraphics[width=1.0\linewidth, clip]{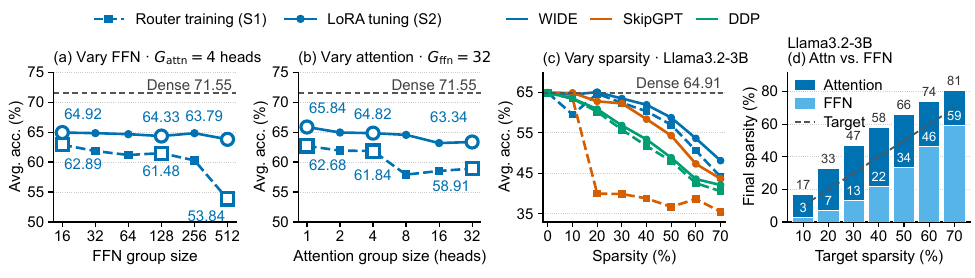}
    \caption{Effects of pruning group size and target sparsity. (a--b) \med~on Llama3.1-8B at 50\% sparsity, varying FFN group size with $G_{\text{attn}}=4$ heads, or attention group size with $G=32$. (c) \med~($G=32$), DDP, and SkipGPT's performance on Llama3.2-3B across target sparsity. (d) Final sparsity distribution of \med~($G=32$) during Llama3.2-3B's training.}
    \label{fig:group_and_sparsity}
\end{figure}

\paragraph{Performance across Group Size and Sparsity}
For \med, group size $G$ and target sparsity govern the quality--speed trade-off. $G_{\text{ffn}}$ spans $[1,D^\prime]$, from row-wise to full-FFN skipping, while $G_{\text{attn}}$ typically uses $\frac{H_q}{H_k}\times d$ but can vary over $[1,H_q]\times d$, up to SkipGPT's granularity. As shown in Fig.~\ref{fig:group_and_sparsity} (a--b), router training favors smaller FFN groups: increasing the group size to 512 reduces accuracy by 9.05 points, also helping explain SkipGPT's weak stage-1 results. After LoRA tuning, sensitivity to group size drops substantially, with only 2.50 and 1.13-point accuracy reductions for attention and FFN from the finest to coarsest granularity, respectively. For target sparsity (Fig.~\ref{fig:group_and_sparsity} (c--d)), \med's accuracy declines steadily but remains above SkipGPT and DDP, with router-trained \med~outperforming LoRA-tuned SkipGPT at most sparsity levels. In terms of sparsity allocation, \med~follows a trend similar to SkipGPT by assigning more sparsity budget to attention, while the attention-to-FFN sparsity ratio gradually decreases from 5.9:1 at 10\% target sparsity to 1.3:1 at 70\%.
\begin{figure}[!tb]
    \centering
    \vspace{0pt}
    \includegraphics[width=1.0\linewidth, clip]{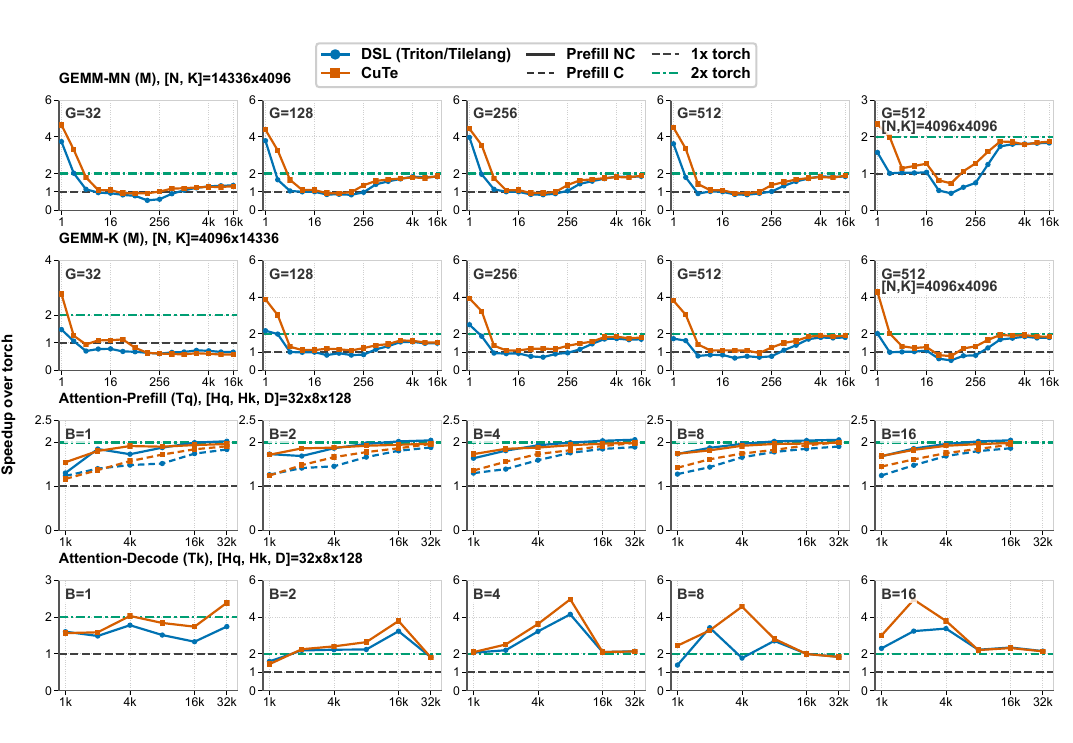}
    \caption{Kernel-level speedups of the four types of \med~kernels on Llama3.1-8B shapes, with a random 50\% sparsity mask. \texttt{NC} and \texttt{C} denote non-causal and causal attention kernels. The dense baseline is set to PyTorch cuBLAS for GEMM, and SDPA (FlashAttention/FlashDecoding backend) for Attention. Overhead from workspace initialization is also taken into account.}
    \label{fig:kernel_bench}
\end{figure}
\begin{figure}[!tb]
    \centering
    \includegraphics[width=1.0\linewidth]{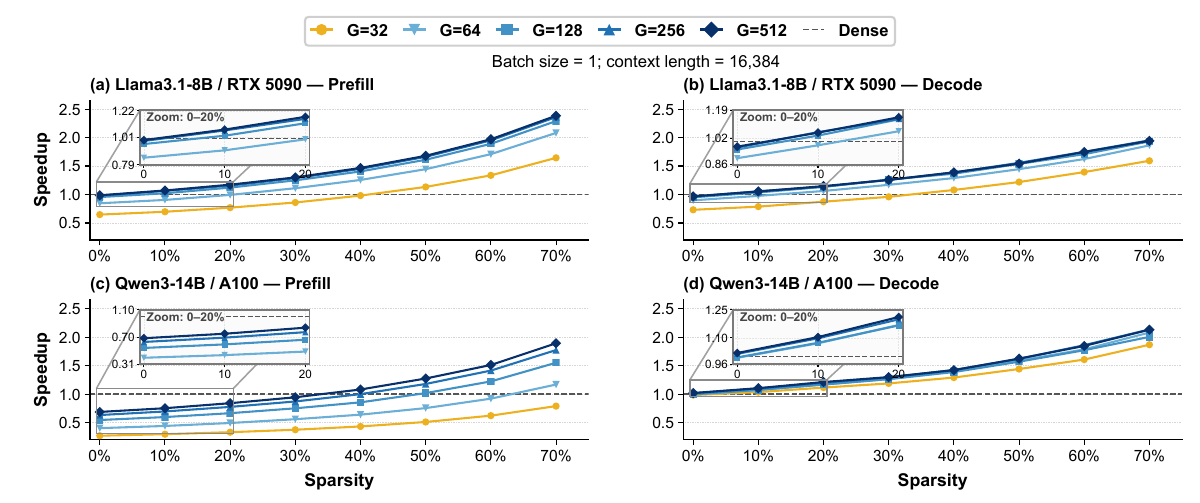}
    \caption{\med's end-to-end speedup over the dense baseline across models, sparsity, and $G$.}
    \label{fig:end_to_end_speedup}
\end{figure}

\paragraph{Kernel-level Speedup}
As shown above, \med's quality is insensitive to group size after LoRA tuning, favoring larger $G$ that provides more tiling options and improves kernel pipeline efficiency. The kernel benchmarks in Fig.~\ref{fig:kernel_bench}, using random 50\% routing masks, show that GEMM-K approaches the ideal speedup over dense kernels only when $G>128$, as smaller K-axis groups provide insufficient mainloop work to amortize pipeline overhead. GEMM-MN likewise requires $G\geqslant128$ to enable flexible tiling. Head-wise attention kernels are less sensitive to $G$, although GQA-packing decoding still requires GQA-aligned groups (Appendix~\ref{sec:Additional Experimental Results}, Fig.~\ref{fig:attention_speedup_vary_g}). In addition, pruned GEMM provides limited acceleration when $BM\geqslant M$ under fully random masks, since the probability of a tile being fully inactive is close to zero in such scenario, leaving limited opportunities for CTA-level skipping.

\begin{figure}[!tb]
    \centering
    \begin{minipage}[t]{0.48\linewidth}
    \includegraphics[width=1.0\linewidth]{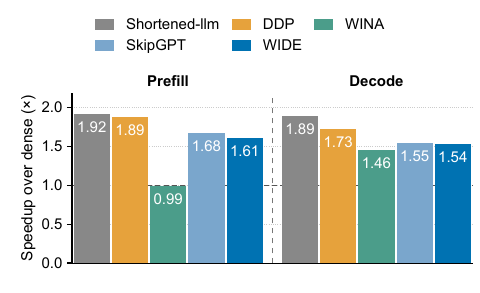}
    \caption{50\% sparsity's end-to-end speedup over the Llama3.1-8B dense baseline across pruning methods.}
    \label{fig:baseline_e2e}
    \end{minipage}\hfill
    \begin{minipage}[t]{0.48\linewidth}
    \includegraphics[width=1.0\linewidth]{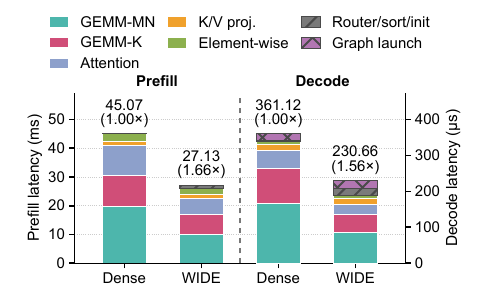}
    \caption{Layer-wise latency breakdown on Llama3.1-8B at 50\% sparsity.}
    \label{fig:latency_breakdown}
    \end{minipage}
\end{figure}
\paragraph{End-to-end Speedup}
Building on the kernel-level results, we further evaluate end-to-end acceleration across different hardware and models under varying sparsity levels and FFN group sizes, with input shape fixed at $B=1$, $T=16{,}384$. As shown in Fig.~\ref{fig:end_to_end_speedup}, the best setting on sm120 with $G\leqslant 128$ retains 0.98x of dense throughput, while 50\% sparsity increases the speedup to 1.68x for prefill and 1.55x for decoding. On sm80, which lacks support for vectorized atomic operations, \med~cannot sustain the same prefill gains, as the pruned GEMM, especially GEMM-K, rely heavily on split-K parallelism and atomic bandwidth. Nevertheless, \med~still delivers up to a 1.62$\times$ speedup in decoding.

Beyond the acceleration landscape of \med~itself, Fig.~\ref{fig:baseline_e2e} further compares the end-to-end speedups of different pruning schemes under the same input shapes. WINA uses the sparse GEMV kernel from TEAL~\citep{liu2024TEAL}, while SkipGPT uses the same kernels as \med. Throughout the evaluation, \med~achieves acceleration comparable to SkipGPT. WINA delivers practical speedup for single-request decoding, but is not compatible with prefill workloads. Overall, \med~provides a better balance between practical acceleration and routing flexibility.

After evaluating the end-to-end acceleration potential of \med~, we further analyze how close it comes to the theoretical speedup bound. In the averaged forward trace of a Llama3.1-8B layer at 50\% sparsity with $G=128$ (Fig.~\ref{fig:latency_breakdown}), the pruned path reaches $1.82\times$ speedup for prefill and $1.92\times$ for decoding, close to the ideal bound, while the remaining work accounts for about 16.7\% and 29.1\% of prefill and decoding time, respectively. These overheads come from the additional computation of routing-based pruning, workspace initialization in the optimized kernels, and fixed costs from unpruned operations, e.g., element-wise operators, explaining why the overall speedup remains below the theoretical bound.

\section{Conclusion}
In this work, we identify the potential limitation for current naive token-wise dynamic pruning pipeline under real-world inference system, and present a finer-grained group-based structured dynamic width pruning method with unified acceleration design for routing-based dynamic structural pruning. The presented \med~uses lightweight bottleneck routers and differentiable masks to select GQA head groups and FFN channel groups for each token, pushing structured pruning below the granularity of layers and sublayers. On the system side, \med~formulates mask-based dynamic pruning as two steps: mask reordering and progressive intra-kernel skipping. Across the evaluations, \med~consistently beats existing static and dynamic structured pruning methods in overall quality retention, while effectively translating sparsity into practical speedups under both prefill and decode workloads. These results demonstrate the potential of \med~for adaptation and optimization across broader model families and inference systems.

\bibliographystyle{plainnat}
\bibliography{cite}

\newpage
\appendix

\section{The Cost for Naive Gather-Scatter Implementations}
\label{apdx:memory-accounting}

In this section, we first provide a narrow theoretical memory accounting for naive gather-scatter implementations of M-axis, N-axis, and K-axis dynamic pruning. We count the logical data-element transfers required to materialize and consume the explicit tensors in the PyTorch-style implementations below. Routing masks, indices, offsets, and their construction, storage, and traffic are excluded, as are hardware-specific cache effects and framework workspace. We assume that \(\mathbf{A}\), \(\mathbf{B}\), and \(\mathbf{C}\) use the same data type, so byte traffic is obtained by multiplying \(T_M\), \(T_N\), or \(T_K\) by the data-type size. Consider a dense GEMM
\begin{align}
    \textbf{C} = \textbf{A}\textbf{B}^\top,\qquad
    \textbf{A}\in\mathbb{R}^{M\times K},\quad
    \textbf{B}\in\mathbb{R}^{N\times K},\quad
    \textbf{C}\in\mathbb{R}^{M\times N}
\end{align}
The ideal dense traffic is
\begin{align}
    T_{\text{dense}} \approx MK + NK + MN
\end{align}
which corresponds to reading \textbf{A}, reading \textbf{B}, and writing \textbf{C}.

\paragraph{Naive M-axis pruning}
M-axis pruning (i.e., dynamic depth pruning) selects a subset of token rows for the current operator. Let \(M'=pM\) be the number of active rows, where \(p\) is the active ratio, and let \(I_M\in\mathbb{Z}^{M'}\) denote the active row indices. A naive PyTorch-style implementation gathers active rows, computes the compact GEMM, and scatters the result back like:
\begin{lstlisting}[style=pytorch]
# A: [M, K]
# B: [N, K]
# I_M: [M']

A_sub = A[I_M, :]                 # [M', K]
C_sub = A_sub @ B.T               # [M', N]

C = torch.zeros(M, N, device=A.device, dtype=A.dtype)
C[I_M, :] = C_sub                 # scatter back to dense output
\end{lstlisting}
When ignoring index traffic, the total memory traffic is
\begin{align}
    T_M
    &\approx
    3M'K + NK + 3M'N + MN \notag\\
    &= 3pMK + NK + 3pMN + MN
\end{align}
The terms \(3M'K\) and \(3M'N\) come from materializing and rereading $\textbf{A}_{\text{sub}}$ and $\textbf{C}_{\text{sub}}$, respectively, while \(MN\) comes from initializing the dense output tensor. The additional storage, excluding the final dense output \textbf{C}, is
\begin{align}
    S_M
    &\approx M'K + M'N \notag\\
    &= pMK + pMN
\end{align}

\paragraph{Naive N-axis pruning.}
For N-axis pruning, different \(N\)-axis groups are selected for each token. We divide the \(N\)-axis into groups of size \(G\) and assume that each token selects \(R\) groups in this baseline. The active \(N\)-axis extent is
\begin{align}
    N' = RG = pN
\end{align}
where \(p\) denotes the active ratio along the \(N\)-axis. Let \(I_N\in\mathbb{Z}^{M\times R}\)  denote the selected group indices for each token. The fully vectorized PyTorch-style implementation below require to expand the group indices and materializes token-specific subsets of \(\mathbf{B}\):
\begin{lstlisting}[style=pytorch]
# A: [M, K]
# B: [N, K]
# I_N: [M, R], group indices
# G: group size
# N_prime = R * G

offset = torch.arange(G, device=A.device)              # [G]

J_N = I_N[..., None] * G + offset                      # [M, R, G]
J_N = J_N.reshape(M, R * G)                            # [M, N']

B_sub = B[J_N, :]                                      # [M, N', K]

C_sub = torch.einsum("mk,mnk->mn", A, B_sub)           # [M, N']

C = torch.zeros(M, N, device=A.device, dtype=A.dtype)
C.scatter_(dim=1, index=J_N, src=C_sub)                # [M, N]
\end{lstlisting}
Ignoring index traffic and the temporary offset data, the overall memory traffic is
\begin{align}
    T_N
    &\approx
    MK + 3MN'K + 3MN' + MN \notag\\
    &= MK + 3pMNK + 3pMN + MN
\end{align}
The dominant term \(3MN'K\) comes from reading selected rows of \textbf{B}, writing the gathered tensor $\textbf{B}_{\text{sub}}$, and reading $\textbf{B}_{\text{sub}}$ again during token-wise dot products. Unlike dense GEMM, where \textbf{B} is shared across all \(M\) rows and read once as \(NK\), the naive N-axis implementation replicates selected weight groups for each token, with additional storage in
\begin{align}
    S_N
    &\approx
    MN'K + MN' \notag\\
    &= pMNK + pMN
\end{align}
which is dominated by the per-token gathered weight tensor $\textbf{B}_{\text{sub}}\in\mathbb{R}^{M\times N'\times K}$, where the remaining \(MN'\) entries belong to $\textbf{C}_{\text{sub}}$.

\paragraph{Naive K-axis pruning.}
For attention output and FFN down projections, K-axis pruning is applied along the reduction dimension \(K\). We divide the \(K\)-axis into groups of size \(G\) and assume that each token selects \(R\) groups. The active (K)-axis extent is
\begin{align}
    K' = RG = pK
\end{align}
where \(p\) denotes the active ratio along the \(K\)-axis. Let \(I_K\in\mathbb{Z}^{M\times R}\) denote the selected \(K\)-group indices for each token. A token-specific PyTorch-style implementation can be written as:
\begin{lstlisting}[style=pytorch]
# A: [M, K]
# B: [N, K]
# I_K: [M, R], group indices
# G: group size
# K_prime = R * G

offset = torch.arange(G, device=A.device)              # [G]

J_K = I_K[..., None] * G + offset                      # [M, R, G]
J_K = J_K.reshape(M, R * G)                            # [M, K']

A_sub = torch.gather(A, dim=1, index=J_K)              # [M, K']
B_sub = B[:, J_K].permute(1, 0, 2)                     # [M, N, K']

C = torch.einsum("mk,mnk->mn", A_sub, B_sub)           # [M, N]
\end{lstlisting}
Ignoring index traffic and the small offset tensor, the total memory traffic is
\begin{align}
    T_K
    &\approx
    3MK' + 3MNK' + MN \notag\\
    &= 3pMK + 3pMNK + MN
\end{align}
The term \(3MK'\) comes from reading selected entries of \(\mathbf{A}\), writing \(\mathbf{A}_{\text{sub}}\), and reading \(\mathbf{A}_{\text{sub}}\) again during the token-wise dot product. The dominant term \(3MNK'\) comes from reading selected entries of \(\mathbf{B}\), writing the token-specific tensor \(\mathbf{B}_{\text{sub}}\), and reading \(\mathbf{B}_{\text{sub}}\) again during the einsum. The additional materialized data-tensor storage is
\begin{align}
    S_K
    &\approx
    MNK' + MK' \notag\\
    &= pMNK + pMK
\end{align}
which is dominated by the token-specific gathered weight tensor \(\mathbf{B}_{\text{sub}}\in\mathbb{R}^{M\times N\times K'}\), where the remaining \(MK'\) entries belong to \(\mathbf{A}_{\text{sub}}\).

These estimates highlight why the fully vectorized N-axis and K-axis implementations above are especially inefficient: they materialize token-specific subsets of \textbf{B}, incurring $\mathcal{O}(pMNK)$ intermediate storage and destroying the regular weight reuse of dense GEMM, which highlights the role of customized fused kernels. The results are summarized in Table \ref{tab:naive-overhead-summary}.

\begin{table}[htbp]
\centering
\small
\resizebox{\linewidth}{!}{
    \begin{tabular}{lcc}
    \toprule
    Pattern & Naive data-tensor traffic & Extra intermediate storage \\
    \midrule
    Dense GEMM &
    \(MK + NK + MN\) &
    -- \\
    Dynamic pruning (M-axis) &
    \(3pMK + NK + 3pMN + MN\) &
    \(pMK + pMN\) \\
    Dynamic pruning (N-axis) &
    \(MK + 3pMNK + 3pMN + MN\) &
    \(pMNK + pMN\) \\
    Dynamic pruning (K-axis) &
    \(3pMK + 3pMNK + MN\) &
    \(pMNK + pMK\) \\
    \bottomrule
    \end{tabular}
}
\caption{Theoretical data-tensor traffic and additional materialized intermediate storage of naive gather-scatter implementations. Routing masks, indices, and offsets are excluded, and the storage calculation also excludes the final dense output \textbf{C}.}
\label{tab:naive-overhead-summary}
\end{table}

\begin{figure}[htbp]
    \centering
    \includegraphics[width=1.0\linewidth]{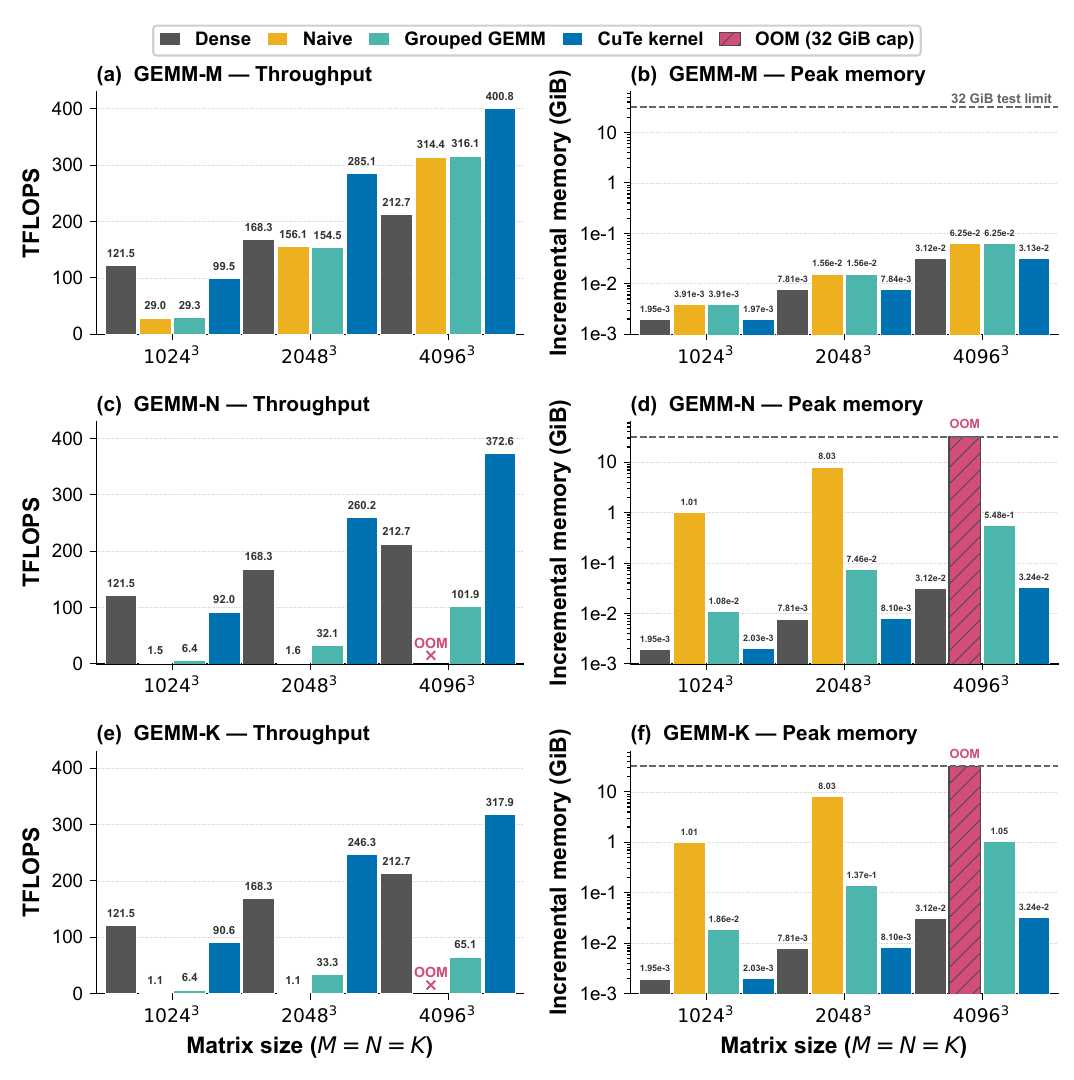}
    \caption{Throughput and memory overhead benchmarks for M, N, and K-axis GEMM pruning based on dense GEMM baseline, naive gather-scatter kernel, grouped-GEMM kernel, and fused CuTe kernel at 50\% sparsity, with $G=128$ for width pruning.}
    \label{fig:naive_vs_cute_gemm}
\end{figure}

Building on the theoretical analysis, we further benchmark the throughput and peak memory usage of the dense GEMM baseline, naive gather-scatter kernel, and optimized CuTe GEMM kernel for pruning along the M, N, and K axes. Since the naive implementation can be adapted to grouped-GEMM by mask-reordering preprocessing, we also introduce it as a stronger baseline. The results are shown in Figure~\ref{fig:naive_vs_cute_gemm}. For the coarsest-grained M-axis pruning setting, the explicit gather-scatter kernel still delivers a meaningful speedup but incurs an additional memory overhead of roughly 2x. For the finer-grained GEMM-N and GEMM-K settings, however, the naive implementation is constrained by the massive additional memory traffic, achieving only about 1.0 TFLOPS and requiring roughly 1,000x the memory footprint compared to the optimized kernels, while the grouped-GEMM kernel still requires up to 34x the incremental memory of dense GEMM and only achieves 6$\sim$102 TFLOPS, failing to reach the dense GEMM baseline. In contrast, the fused CuTe kernel incurs virtually no additional memory-storage overhead and can already deliver speedups for matrices of size $2048^3$.
\FloatBarrier

\section{CuTe Kernel Details}
\label{apdx:cute-kernel-details}
At a high level, all three CuTe kernels follow the same execution path as the framework in the main text: (1) standalone mask-reordering preprocessing, (2) hardware-agnostic block-level skipping at stage $\ell=0$, (3) a dense-style tiled pipeline with optional hardware-dependent intra-block skipping during operand gathering ($\ell=1$, corresponding to $\mathbf{A}$ loading in GEMM and $\mathbf{Q}$ loading in attention) and MMA execution ($\ell=2$), and (4) a scatter epilogue. The $\ell=1$ and $\ell=2$ predicates are enabled only when specified by the architecture-specific kernel configuration.

Herein, we first describe the metadata shared by all three kernels. A \emph{routing column} is $\mathcal{M}_{:,g}$ after flattening the batch and token dimensions for GEMM, and $\mathcal{M}_{b,:,g}$ for FlashAttention. As shown in Algorithm~\ref{alg:route-preprocess}, each routing column is sorted independently on the GPU. The resulting index $\mathcal{I}_{j,g}$ maps logical row $j$ to its physical row. Activation tensors remain in their original layout and are gathered/scattered directly through the reordered indices. Consecutive operators sharing the same routing decision can reuse this metadata, while zero initialization still remains necessary for skipped blocks and atomic-add epilogues.

\begin{algorithm}[htbp]
\caption{Unified mask reordering}
\label{alg:route-preprocess}
\begin{algorithmic}[1]
\Require Binary routing mask $\mathcal{M}$
\Ensure Reordered mask $\widetilde{\mathcal{M}}$ and original row indices $\mathcal{I}$
\ForAll{groups $g$ (and batch elements $b$ for attention) in parallel}
    \State $(\widetilde{\mathcal{M}}_{:,g},\mathcal{I}_{:,g})
           \gets\operatorname{SortDesc}(\mathcal{M}_{:,g})$
           \Comment{batch index omitted}
\EndFor
\State \Return $(\widetilde{\mathcal{M}},\mathcal{I})$
\end{algorithmic}
\end{algorithm}
\FloatBarrier

For a sorted CTA row tile beginning at $m_0$, the block-level predicate is simply $s^{(0)}=\widetilde{\mathcal{M}}_{m_0,g}$. Since the column is monotone, $s^{(0)}=0$ can prove that the whole CTA and every later CTA for this group are inactive. Conversely, if a full tile's last entry is active, the kernel takes an all-active fast path and avoids loading per-row masks. For contiguous row blocks, the predicates simplify to
\begin{align}
    s^{(\ell)}_{m,g}=\widetilde{\mathcal{M}}_{mS_\ell,g},
    \qquad mS_\ell<M
\end{align}
The $\ell=1$ predicate is instantiated per thread or per warp according to the CuTe copy atom. The $\ell=2$ predicate is reduced with a warp vote instruction and therefore remains uniform across all lanes issuing the same tensor-core instruction. The two finer skipping levels are independently configurable, disabling either level causes execute all its in-bounds units.

In the pseudocode below, $s^{(1)}$ and $s^{(2)}$ collect the predicates within the admitted tile, and $\mathcal{I}$ directly supplies the original row indices. We omit CuTe layout construction, register-fragment permutations, and architecture-specific pipeline stages and barriers, with four primitives shared by all pseudo kernels below: \textsc{Gather} loads reordered activation rows admitted by $s^{(1)}$; \textsc{Load} stages a dense weight or KV tile; \textsc{MaskedMMA} accumulates row fragments admitted by $s^{(2)}$; and \textsc{Scatter} writes only rows with $\widetilde{\mathcal{M}}=1$, with configurable behavior between direct stores (\texttt{store}) and atomic addition (\texttt{add}). Additionally, the \textsc{MaskedMMA} in sm80 and sm120 also denote to skip the explicit shared-to-register transfer required by their tensor core.

\paragraph{Implementation of GEMM-MN}
GEMM-MN is the shared kernel for M-axis and N-axis pruning. Its N-axis path applies routing to the attention $\textbf{W}_q$ and FFN $\textbf{W}_{\text{up}}$/$\textbf{W}_{\text{gate}}$, while M-axis dynamic depth pruning is the $N_G=1$ special case of the same execution path. Let $\mathbf{C}=\mathbf{A}\mathbf{B}^{\top}$, with $\mathbf{A}\in\mathbb{R}^{M\times K}$ and $\mathbf{B}\in\mathbb{R}^{N\times K}$, and, for N-axis pruning, partition $N$ into groups of width $G$, with $BN\mid G$. Each CTA owns one $(BM,BN)$ output tile, whose $N$ coordinate selects a routing group $g$, where M-axis pruning uses $g=0$ for every output tile. Algorithm~\ref{alg:gemm-mn} makes the CTA admission, predicated computation, and scatter epilogue explicit.

\begin{algorithm}[htbp]
\caption{GEMM-MN for output tile $(m_0,n_0)$}
\label{alg:gemm-mn}
\begin{algorithmic}[1]
\Require $\mathbf{A},\mathbf{B}$, zero-initialized $\mathbf{C}$,
         $(\widetilde{\mathcal{M}},\mathcal{I})$, tile sizes $(BM,BN,BK)$, assigned $K$ range
\State $g\gets 0$ for M-axis pruning; otherwise $g\gets\lfloor n_0/G\rfloor$
\If{$\widetilde{\mathcal{M}}_{m_0,g}=0$}
    \State \Return \Comment{$\ell=0$: skip the output tile}
\EndIf
\State Read $s^{(1)},s^{(2)}$ and $\mathcal{I}_{m_0:m_0+BM,g}$
\State $\mathbf{C}_{\mathrm{acc}}\gets\mathbf{0}$
\For{each $BK$-tile beginning at $k$ in the assigned $K$ range}
    \State $\mathbf{A}_{\mathrm{tile}}\gets\Call{Gather}{\mathbf{A}[\mathcal{I}_{m_0:m_0+BM,g},k:k+BK],s^{(1)}}$
    \State $\mathbf{B}_{\mathrm{tile}}\gets\Call{Load}{\mathbf{B}[n_0:n_0+BN,k:k+BK]}$
    \State $\mathbf{C}_{\mathrm{acc}}\gets\Call{MaskedMMA}
           {\mathbf{A}_{\mathrm{tile}},\mathbf{B}_{\mathrm{tile}}^{\top},\mathbf{C}_{\mathrm{acc}},s^{(2)}}$
\EndFor
\State $\texttt{mode}\gets\texttt{add}$ for split-$K$; otherwise $\texttt{mode}\gets\texttt{store}$
\State \Call{Scatter}{$\mathbf{C}_{\mathrm{acc}}$, $\mathbf{C}[\mathcal{I}_{m_0:m_0+BM,g},n_0:n_0+BN]$, \texttt{mode}}
\end{algorithmic}
\end{algorithm}
\FloatBarrier

It is clear that the standard dense GEMM pipeline still remaining, since each logical $K$ tile still commits exactly one asynchronous-copy group, the sparse predicates suppress only $\mathbf{A}$ load packets, with dense $\mathbf{B}$ copy keeps the producer stage well-defined. Without split-$K$, the assigned range is $[0,K)$ and each output entry has one writer. Split-$K$ variants partition this range across CTAs and atomically reduce the scattered partial outputs into the zero-initialized buffer. The GEMV variant can be dispatched when $M\leqslant 4$.

\paragraph{Implementation of GEMM-K}
GEMM-K implements the K-axis path for the attention $\mathbf{W}_{\text{o}}$ and FFN $\mathbf{W}_{\text{down}}$. Here the routed dimension is the reduction dimension, with $N_G=K/G$:
\begin{align}
    \mathbf{C}_{m,n}
    =
    \sum_{g=0}^{N_G-1}\mathcal{M}_{m,g}
    \sum_{k=gG}^{(g+1)G-1}\mathbf{A}_{m,k}\mathbf{B}_{n,k}
\end{align}
Since different groups have different row permutations, each active group computes a partial output and scatters it with an atomic reduction, as shown in Algorithm~\ref{alg:gemm-k}. For each $(m_0,n_0)$, reduction groups are partitioned among CTAs, with $BK\mid G$.

\begin{algorithm}[htbp]
\caption{GEMM-K for row/output tile $(m_0,n_0)$ and assigned reduction groups}
\label{alg:gemm-k}
\begin{algorithmic}[1]
\Require $\mathbf{A},\mathbf{B}$, zero-initialized $\mathbf{C}$,
         $(\widetilde{\mathcal{M}},\mathcal{I})$, tile sizes $(BM,BN,BK)$
\For{each reduction group $g$ assigned to this CTA}
    \If{$\widetilde{\mathcal{M}}_{m_0,g}=1$} \Comment{$\ell=0$: admit this group}
        \State Read $s^{(1)},s^{(2)}$ and $\mathcal{I}_{m_0:m_0+BM,g}$
        \State $\mathbf{C}_{\mathrm{acc}}\gets\mathbf{0}$
        \For{each $BK$-tile beginning at $k$ in $[gG,(g+1)G)$}
            \State $\mathbf{A}_{\mathrm{tile}}\gets\Call{Gather}{\mathbf{A}[\mathcal{I}_{m_0:m_0+BM,g},k:k+BK],s^{(1)}}$
            \State $\mathbf{B}_{\mathrm{tile}}\gets\Call{Load}{\mathbf{B}[n_0:n_0+BN,k:k+BK]}$
            \State $\mathbf{C}_{\mathrm{acc}}\gets\Call{MaskedMMA}
                   {\mathbf{A}_{\mathrm{tile}},\mathbf{B}_{\mathrm{tile}}^{\top},\mathbf{C}_{\mathrm{acc}},s^{(2)}}$
        \EndFor
        \State \Call{Scatter}{$\mathbf{C}_{\mathrm{acc}}$, $\mathbf{C}[\mathcal{I}_{m_0:m_0+BM,g},n_0:n_0+BN]$, \texttt{add}}
    \EndIf
\EndFor
\end{algorithmic}
\end{algorithm}
\FloatBarrier

Unlike GEMM-MN, GEMM-K scopes its pipeline to one active $K$-group. The producer--consumer ring and accumulator reset at each group boundary, while an inactive group skips its copies, synchronizations, MMA, and epilogue work entirely. This group-local design sacrifices prefetching across adjacent $K$-groups, but allows $\ell=0$ block-level skipping to bypass the entire inner pipeline. Advancing a shared ring over skipped groups would otherwise leave unmatched wait/commit operations and increase barrier-management overhead. It also creates substantial atomic-reduction contention because around $\lceil K/G\rceil$ groups may scatter partial outputs to the same entries of $\mathbf{C}$. We mitigate this contention with an outer loop that packs multiple $K$-groups into one CTA and executes them sequentially, thus reducing concurrent writebacks without changing the number of group contributions. For small $M$, the GEMV variant is dispatched according to both $M$ and $G$.

On sm90+, TMA also supports asynchronous tensor reduction from shared to global memory through \texttt{cp.reduce.async.bulk.tensor}. This enables atomic writeback from shared memory, allowing a warp-specialized pipeline to overlap the atomic epilogue with the mainloop and potentially improve GEMM-K performance for small $G$. This optimization is less suitable for sm120: persistently storing $\mathbf{C}$ in shared memory consumes too much of the limited shared-memory capacity and substantially reduces occupancy. We therefore leave it as future work.

\paragraph{Implementation of Attention}
As for prefill attention, a CTA owns a query tile at batch $b$ and head $h_q$. The query head determines both its KV head for GQA and routing group $g$. Algorithm~\ref{alg:attention} follows the tiled online-softmax organization of FlashAttention~\citep{FlashAttention2}, while fusing the same three predicates into the query-side work. Its path is query-tile admission, $\mathbf{Q}$ prologue, dense $\mathbf{K}/\mathbf{V}$ pipelined mainloop, and scatter epilogue. Here \textsc{OnlineSoftmax} updates the running maximum $\mathbf{m}$ and normalization sum $\mathbf{z}$, rescales $\mathbf{O}_{\mathrm{acc}}$, and returns unnormalized weights $\mathbf{P}$ for active rows. A row with no valid keys in a tile uses $\mathbf{P}=\mathbf{0}$ and leaves its running state unchanged.

Using the head size $d$ and group size $G$ from the main text, the KV head and routing group are
\begin{align}
    h_k=\left\lfloor\frac{h_q}{H_q/H_k}\right\rfloor,
    \qquad
    g=\left\lfloor\frac{h_q}{G/d}\right\rfloor
\end{align}
For the broader causal attention scenario, since tokens in a gathered query tile have different real position, the causal masking admits key position $j$ only if $j\leqslant\mathcal{I}_{b,m,g}$. Therefore, Algorithm~\ref{alg:attention} uses a CTA-wide \textsc{ReduceMax} over active query positions to allocate the maximum position in this query tile.

\begin{algorithm}[htbp]
\caption{Routed causal prefill attention for query tile $(b,h_q,m_0)$}
\label{alg:attention}
\begin{algorithmic}[1]
\Require $\mathbf{Q},\mathbf{K},\mathbf{V}$, zero-initialized $\mathbf{O}$,
         $(\widetilde{\mathcal{M}},\mathcal{I})$, tile sizes $(BM,BN)$
\State Map $h_q$ to KV head $h_k$ and routing group $g$
\If{$\widetilde{\mathcal{M}}_{b,m_0,g}=0$}
    \State \Return \Comment{$\ell=0$: skip the query tile}
\EndIf
\State Read $s^{(1)},s^{(2)}$ and $\mathcal{I}_{b,m_0:m_0+BM,g}$
\State $j_{\max}\gets\min\!\left(T_k,1+\Call{ReduceMax}{\mathcal{I}_{b,m_0:m_0+BM,g}}\right)$
\State $\mathbf{Q}_{\mathrm{tile}}\gets\Call{Gather}
       {\mathbf{Q}[b,\mathcal{I}_{b,m_0:m_0+BM,g},h_q,:],s^{(1)}}$
\State $(\mathbf{m},\mathbf{z},\mathbf{O}_{\mathrm{acc}})
       \gets(-\infty,\mathbf{0},\mathbf{0})$
\For{each aligned key tile $[j,j+BN)$ intersecting $[0,j_{\max})$}
    \State $\mathbf{K}_{\mathrm{tile}}\gets\Call{Load}{\mathbf{K}[b,j:j+BN,h_k,:]}$
    \State $\mathbf{V}_{\mathrm{tile}}\gets\Call{Load}{\mathbf{V}[b,j:j+BN,h_k,:]}$
    \State $\mathbf{S}\gets\Call{MaskedMMA}
           {\mathbf{Q}_{\mathrm{tile}},\mathbf{K}_{\mathrm{tile}}^{\top},\mathbf{0},s^{(2)}}/\sqrt{d}$
    \State Apply causal masking using $\mathcal{I}$; set inactive rows and invalid keys of $\mathbf{S}$ to $-\infty$
    \State $(\mathbf{m},\mathbf{z},\mathbf{O}_{\mathrm{acc}},\mathbf{P})
           \gets\Call{OnlineSoftmax}{\mathbf{S},\mathbf{m},\mathbf{z},\mathbf{O}_{\mathrm{acc}}}$
    \State $\mathbf{O}_{\mathrm{acc}}\gets\Call{MaskedMMA}
           {\mathbf{P},\mathbf{V}_{\mathrm{tile}},\mathbf{O}_{\mathrm{acc}},s^{(2)}}$
\EndFor
\State Normalize active rows: $\mathbf{O}_{\mathrm{acc}}\gets\mathbf{O}_{\mathrm{acc}}/\mathbf{z}$
\State \Call{Scatter}{$\mathbf{O}_{\mathrm{acc}}$, $\mathbf{O}[b,\mathcal{I}_{b,m_0:m_0+BM,g},h_q,:]$, \texttt{store}}
\end{algorithmic}
\end{algorithm}
\FloatBarrier

Once a CTA is admitted, $K$ and $V$ remain dense and their two producer--consumer pipelines advance once per logical key tile. Row-level predicates never suppress a barrier arrival or stage transition. They only suppress the gathered $Q$ packets and the QK/PV MMA fragments, preserving barrier phase agreement across the CTA. For decoding ($T_q=1$) phase, routing can be checked directly without mask reordering, while the split online-softmax pipeline follows the same abstraction.
\FloatBarrier

\section{Implementation Details for Training and Evaluation}\label{sec:Implementation Details for Training and Evaluation}
Since different model families have their own training data distributions, we use separate calibration datasets for the Llama3 and Qwen3 families\footnote{https://huggingface.co/datasets/ZengXiangyu/RedPajama-Data-1T-Sample}\footnote{https://huggingface.co/datasets/BAAI/Infinity-Instruct}. Specifically, we apply a uniform sampling strategy to each full dataset for constructing a calibration set of around 200,000 untruncated samples with tokenized lengths less than the maximum $T_q$. For Infinity-Instruct, we also apply an assistant-only mask during end-to-end training, including S1 in DDP, SkipGPT, and~\med, as well as LoRA fine-tuning for all methods in S2. The learning rate for \med's S1 is selected by grid search over \{2e-3, 1e-3, 5e-4, 2e-4\}, where SkipGPT fails to converge on Qwen3-14B at its default learning rate of 2e-3, so 5e-4 is also applied to SkipGPT in Qwen3-14B training. Both training stages run on 4 NVIDIA A100-SXM4-40G GPUs with PyTorch FSDP2. We further apply gradient checkpointing separately to the attention and FFN modules to reduce training memory usage without replaying the router. During training, to ensure fair comparison, all methods use the same subset for calibration and LoRA tuning. The calibration stage follows each method's original training budget, while the LoRA stage uses the same settings as \med. For SkipGPT, we additionally increase the router rank $r$ to match the router capacity of \med. More detailed settings are listed in Table~\ref{tbl:detailed_params}.

During evaluation, we test all tasks using lm-evaluation-harness~\citep{lm-eval-harness}, with a maximum sequence length of 4,096 and a generation limit of 1,024 tokens, except for IFEval, which retains its default limit of 1,280 tokens. The inference engine is set to native Transformers, enabling a static cache and \texttt{torch.compile} and disabling sampling throughout. For the attention backend, we use torch SDPA to automatically select among the FlashAttention, memory efficient, and math backends, as in the training stages. All tasks except MMLU are evaluated in the zero-shot setting. In addition, for Qwen3-14B, we find that all pruned models achieve better quality in likelihood tasks when using the chat template, while the dense model does not. We therefore apply the chat template to all methods except the dense baseline for Qwen3-14B.

For the experimental environments, all experiments are conducted with CUDA 13.0, PyTorch 2.11, Transformers 5.12.1, and lm-eval 0.4.12. We use Lightning Fabric 2.6.5 for training and SGLang 0.5.16, Triton 3.7.1, and TileLang 0.1.11 for inference kernel implementation and validation. The CuTe kernel libraries are wrapped with the TVM-FFI Just-In-Time (JIT) interface used by SGLang~\citep{SGLang}, combined with an autotuning system to easily switch to the optimal tiling and intra-kernel skipping configuration. Among these DSL implementations, CuTe provides the most fine-grained control and enables intra-block skipping during both $\mathbf{A}$ loading and MMA execution, while the tile-level DSL implementations only support CTA-level skipping due to their coarser control granularity.

During throughput benchmarks, we use random masks with top-k selection to maintain the target sparsity for each kernel, pruned module, and method. Kernel-level benchmarks report the latency of the complete function replay, including device-side kernel execution, host-side operations before and after kernel launches, and other associated latency. The Triton benchmark interface is introduced as the kernel profiling tool, with a repetition window of 200ms for repeated measurements. For end-to-end evaluation, we use a fixed-length static cache to avoid runtime KV-cache allocation, and further exclude the final sampling and detokenization latency. Before measurement, we perform one warmup iteration to complete kernel tuning, then capture a CUDA Graph and replay it 20 times to report the average latency from CUDA event.

\begin{table}[htbp]
    \centering
    \caption{Detailed hyperparameter settings for \med's training stages.}
    \aboverulesep=0ex
    \belowrulesep=0ex
    \renewcommand{\arraystretch}{1.2}
    \begin{tabular}{l|ccc}
        \toprule
        Params. & Llama3.1-8B & Llama3.2-3B & Qwen3-14B \\
        \midrule
        Calibration Data & \multicolumn{2}{c}{RedPajama-1T} & Infinity-Instruct (7M-Core) \\
        Maximum LR (S1) & \multicolumn{2}{c}{2e-3} & 5e-4 \\
        $H_q/H_k$ & 32/8 & 24/8 & 40/8 \\
        $G_{\text{attn}}/d$ (heads) & 4 & 3 & 5 \\
        $G_{\text{ffn}}$ & \multicolumn{3}{c}{\{16, 32, 64, 128, 256, 512\}} \\
        Router $r$ & \multicolumn{3}{c}{32} \\
        $B$ & \multicolumn{3}{c}{16} \\
        Maximum $T_q$ & \multicolumn{3}{c}{4,096} \\
        Initial $\tau$ & \multicolumn{3}{c}{5} \\
        Final $\tau$ & \multicolumn{3}{c}{0.5} \\
        Warmup Steps (S1) & \multicolumn{3}{c}{1,000} \\
        Training Steps (S1) & \multicolumn{3}{c}{10,000} \\
        Warmup Steps (S2) & \multicolumn{3}{c}{1,000} \\
        Training Steps (S2) & \multicolumn{3}{c}{10,000} \\
        Maximum LR (S2) & \multicolumn{3}{c}{2e-4} \\
        LoRA $\alpha$ & \multicolumn{3}{c}{32} \\
        LoRA $r$ & \multicolumn{3}{c}{16} \\
        LoRA dropout & \multicolumn{3}{c}{0.1} \\
        Optimizer & \multicolumn{3}{c}{AdamW (weight decay = 0.01, [$\beta_1$, $\beta_2$] = [0.9, 0.999])} \\
        \bottomrule
    \end{tabular}
    \label{tbl:detailed_params}
\end{table}

\FloatBarrier

\section{Additional Experimental Results}\label{sec:Additional Experimental Results}
\begin{figure}[htbp]
    \centering
    \includegraphics[width=1.0\linewidth, clip]{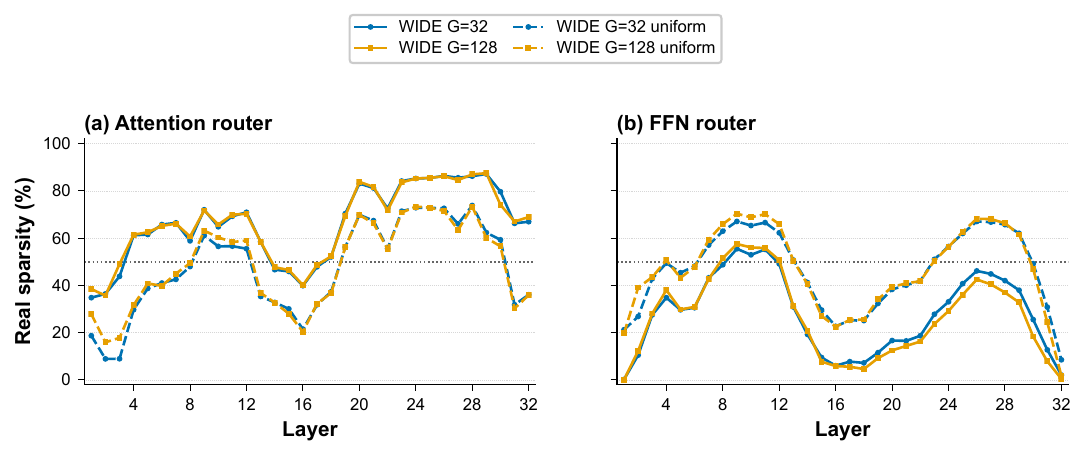}
    \caption{\med's layer-wise sparsity with different $G$ and sparsity allocation.}
    \label{fig:additional_real_sparsity}
\end{figure}

\paragraph{Benchmark over Uniform Sparsity}
We further evaluate \med~with a uniform sparsity constraint that allocates the pruning budget equally between the attention and FFN modules. Specifically, we decompose the sparsity loss into separate attention and FFN terms:
\begin{align}
    \mathcal{L}
    =\mathcal{L}_{\text{LM}}+\frac{\alpha}{2}
    \Bigg(
    \left\lvert SP-\frac{1}{L}\sum_{i=1}^{L}
    \left(1-\frac{\sum\mathcal{M}_{\text{attn}}^i}{\lvert\mathcal{M}_{\text{attn}}^i\rvert}\right)
    \right\rvert
    +
    \left\lvert SP-\frac{1}{L}\sum_{i=1}^{L}
    \left(1-\frac{\sum\mathcal{M}_{\text{ffn}}^i}{\lvert\mathcal{M}_{\text{ffn}}^i\rvert}\right)
    \right\rvert
    \Bigg)
\end{align}
This design constrains the sparsity of the two modules independently rather than matching only their average, keeping their module-wise sparsity levels close to a 1:1 ratio. As shown in Table~\ref{tbl:real_sp_bench}, the uniform constraint causes a larger accuracy drop at larger group sizes, but LoRA recovery restores most of the lost accuracy. After recovery, the uniform variant is within roughly 4\% of the unconstrained \med~variant and still significantly outperforms static structured pruning at the same budget. The layer-wise sparsity profiles in Fig.~\ref{fig:additional_real_sparsity} support the same interpretation: enforcing uniformity between attention and FFN does not alter how sparsity is distributed across layers, it only shifts the profile without changing its shape. Thus, the allocation learned by \med~appears to capture a stable layer-importance pattern induced by the router.

\begin{table}[htbp]
    \centering
    \caption{Downstream performance and the observed average sparsity on Llama3.1-8B with DDP, SkipGPT, and \med~at 50\% target sparsity, where \emph{uniform} denotes imposing a 50\% sparsity target separately on attention and FFN. Sparsity pairs report the sparsity in attention/FFN module.}
    \aboverulesep=0ex
    \belowrulesep=0ex
    \renewcommand{\arraystretch}{1.2}
    \resizebox{1.0\linewidth}{!}{
        \begin{tabular}{l ccccccccc|c}
            \toprule
            Methods & Real Sparsity & WikiText2 & ARC-E & ARC-C & BoolQ & Winogrande & PIQA & OpenbookQA & Hellaswag & Avg. Acc.\\
            \midrule
            DDP & 41.6\%/47.9\% & 21.83 & 55.13 & 33.36 & 65.69 & 60.77 & 69.31 & 30.80 & 56.28 & 53.05 (74.14\%) \\
            \quad -- \emph{LoRA} & 41.6\%/47.9\% & 19.52 & 55.77 & 34.56 & 68.26 & 61.80 & 71.00 & 33.60 & 57.46 & 54.64 (76.36\%) \\
            \midrule
            SkipGPT & 56.7\%/38.7\% & 96.04 & 35.69 & 24.57 & 57.65 & 52.49 & 58.71 & 26.60 & 41.92 & 42.52 (59.42\%)\\
            \quad -- \emph{uniform} & 49.1\%/47.9\% & 104.91 & 33.16 & 24.91 & 48.72 & 50.59 & 60.23 & 27.60 & 31.90 & 39.58 (55.32\%)\\
            \quad -- \emph{LoRA} & 62.7\%/31.6\% & 13.90 & 72.01 & 42.24 & 65.81 & 64.80 & 75.68 & 42.80 & 67.87 & 61.60 (86.10\%)\\
            \quad -- \emph{uniform + LoRA} & 47.6\%/46.5\% & 16.17 & 66.20 & 37.54 & 64.40 & 63.85 & 74.32 & 38.00 & 64.45 & 58.39 (81.61\%)\\
            \midrule
            \med (32) & 63.1\%/30.7\% & 14.15 & 69.74 & 41.55 & 70.31 & 63.69 & 76.22 & 42.20 & 69.17 & 61.84 (86.43\%) \\
            \quad -- \emph{uniform} & 47.8\%/47.1\% & 14.97 & 61.87 & 37.29 & 63.82 & 64.96 & 73.56 & 39.60 & 64.76 & 57.98 (81.03\%) \\
            \quad -- \emph{LoRA} & 66.1\%/27.9\% & 11.99 & 72.90 & 44.20 & 77.22 & 65.98 & 78.40 & 43.40 & 71.66 & 64.82 (90.60\%) \\
            \quad -- \emph{uniform + LoRA} & 47.6\%/46.3\% & 13.62 & 68.39 & 40.78 & 73.43 & 66.85 & 75.90 & 42.40 & 68.83 & 62.36 (87.16\%) \\
            \midrule
            \med (128) & 63.0\%/30.0\% & 12.57 & 66.37 & 39.33 & 69.51 & 67.32 & 76.55 & 42.20 & 69.14 & 61.49 (85.94\%)\\
            \quad -- \emph{uniform} & 48.8\%/48.5\% & 21.58 & 51.64 & 30.89 & 57.13 & 55.01 & 71.55 & 32.60 & 58.17 & 50.99 (71.27\%)\\
            \quad -- \emph{LoRA} & 66.2\%/28.5\% & 11.43 & 71.04 & 41.64 & 75.38 & 67.72 & 78.94 & 43.80 & 71.85 & 64.34 (89.92\%)\\
            \quad -- \emph{uniform + LoRA} & 48.5\%/46.8\% & 13.03 & 68.73 & 39.08 & 73.06 & 66.61 & 76.22 & 40.40 & 67.30 & 61.62 (86.13\%)\\
            \bottomrule
        \end{tabular}
    }
    \label{tbl:real_sp_bench}
\end{table}
\begin{table}[htbp]
    \centering
    \caption{Additional baseline performance under Llama3.1-8B with target sparsity set to 50\%. "S2A2E8" for CMoE denotes splitting the dense FFN into 8 experts, with 2 shared experts and top-2 expert routing. \textbf{A} / \textbf{F} denotes pruning only the attention / FFN module.}
    \aboverulesep=0ex
    \belowrulesep=0ex
    \renewcommand{\arraystretch}{1.2}
    \resizebox{1.0\linewidth}{!}{
        \begin{tabular}{l |c|ccccccc|c}
            \toprule
            \multirow{2}{*}{Methods} & WikiText2 & ARC-E & ARC-C & BoolQ & Winogrande & PIQA & OpenbookQA & Hellaswag & \multirow{2}{*}{Avg. Acc.$\uparrow$}\\
            & (ppl$\downarrow$) & (Acc. Norm.$\uparrow$) & (Acc. Norm.$\uparrow$) & (Acc. $\uparrow$) & (Acc. $\uparrow$) & (Acc. Norm.$\uparrow$) & (Acc. Norm.$\uparrow$) & (Acc. Norm.$\uparrow$) &\\
            \midrule
            \multicolumn{10}{c}{\textbf{Calibration-only (S1)}}\\
            \midrule
            CMoE (S2A2E8)$^{\text{F}}$ & 40.35 & 45.03 & 27.90 & 58.62 & 55.80 & 60.39 & 30.20 & 41.43 & 45.62 (63.76\%) \\
            CMoE (S4A4E16)$^{\text{F}}$ & 40.72 & 44.99 & 28.41 & 63.06 & 55.41 & 62.30 & 30.20 & 41.45 & 46.54 (65.05\%) \\
            HyperPrune & 20.30 & 58.67 & 35.40 & 69.57 & 66.69 & 70.72 & 33.80 & 56.69 & 55.93 (78.17\%) \\
            PolarSparsity$^{\text{A}}$ & 14.03 & 72.52 & 42.41 & 61.04 & 63.85 & 77.58 & 39.80 & 66.64 & 60.54 (84.62\%) \\
            WINA & \underline{9.12} & \underline{77.69} & \underline{51.11} & \textbf{81.74} & \textbf{70.96} & \underline{78.35} & \textbf{43.20} & \underline{76.63} & \underline{68.52 (95.77\%)} \\
            WIDE (32) & 14.15 & 69.74 & 41.55 & 70.31 & 63.69 & 76.22 & 42.20 & 69.17 & 61.84 (86.43\%) \\
            WIDE (128) & 12.57 & 66.37 & 39.33 & 69.51 & 67.32 & 76.55 & 42.20 & 69.14 & 61.49 (85.94\%) \\
            WIDE (128)$^{\text{F}}$ & 15.32 & 63.30 & 38.99 & 66.24 & 64.40 & 73.56 & 39.60 & 64.48 & 58.65 (81.97\%) \\
            WIDE (128)$^{\text{A}}$ & \textbf{8.68} & \textbf{78.57} & \textbf{53.32} & \underline{80.97} & \underline{70.87} & \textbf{80.08} & \underline{43.00} & \textbf{78.57} & \textbf{69.34 (96.91\%)} \\
            \midrule
            \multicolumn{10}{c}{\textbf{LoRA Tuning (S2)}}\\
            \midrule
            CMoE (S4A4E16)$^{\text{F}}$ & 17.26 & 57.95 & 31.48 & 70.21 & 59.12 & 67.14 & 31.60 & 56.44 & 53.42 (74.66\%) \\
            WINA & 9.09 & 77.82 & \underline{51.11} & \underline{79.72} & \underline{70.96} & 79.71 & 43.40 & \underline{77.09} & \underline{68.54 (95.79\%)} \\
            PolarSparsity$^{\text{A}}$ & \underline{8.94} & \underline{78.32} & 50.68 & 77.77 & 67.96 & \textbf{79.98} & \underline{43.80} & 76.72 & 67.89 (94.88\%) \\
            WIDE (32) & 11.99 & 72.90 & 44.20 & 77.22 & 65.98 & 78.40 & 43.40 & 71.66 & 64.82 (90.60\%) \\
            WIDE (128) & 11.43 & 71.04 & 41.64 & 75.38 & 67.72 & 78.94 & \underline{43.80} & 71.85 & 64.34 (89.92\%) \\
            WIDE (128)$^{\text{F}}$ & 12.15 & 71.63 & 41.64 & 78.35 & 64.80 & 76.17 & 41.20 & 70.09 & 63.41 (88.62\%) \\
            WIDE (128)$^{\text{A}}$ & \textbf{8.62} & \textbf{79.46} & \textbf{52.81} & \textbf{81.37} & \textbf{71.58} & \underline{79.97} & \textbf{44.80} & \textbf{78.67} & \textbf{69.80 (97.56\%)} \\
            \bottomrule
        \end{tabular}
    }
    \label{tbl:additional_baseline_bench}
\end{table}

\paragraph{Benchmark over Broader Baselines}
Beyond the static-depth, static-width, and dynamic-depth baselines considered in the main experiments, we broaden our comparison to three additional pruning paradigms: Dense-to-MoE conversion with similar granularity in FFN compared to~\med, semi-structured pruning, and activation-based contextual structured / unstructured pruning at the finest granularity. Specifically, we evaluate CMoE~\citep{pei2026CMoE}, HyperPrune~\citep{sun2025HyperPrune}, PolarSparsity~\citep{shrestha2025Polar}, and WINA~\citep{chen2025WINA} on Llama3.1-8B under a 50\% sparsity budget. We follow the training budget specified in each original paper for calibration, followed by a unified 10k-step LoRA schedule for stage 2. Specifically, PolarSparsity only prunes the attention module due to the non-ReLU activation in Llama3.1-8B, CMoE only prunes the FFN, leaving the full attention module dense, whereas HyperPrune and WINA prune only the linear projections in each layer, while keeping the attention computation itself dense. As shown in Table~\ref{tbl:additional_baseline_bench}, \med~clearly outperforms CMoE and the semi-structured HyperPrune while jointly pruning attention and the FFN, but still trails WINA with its finer pruning granularity.

\begin{figure}[htbp]
    \centering
    \includegraphics[width=1.0\linewidth]{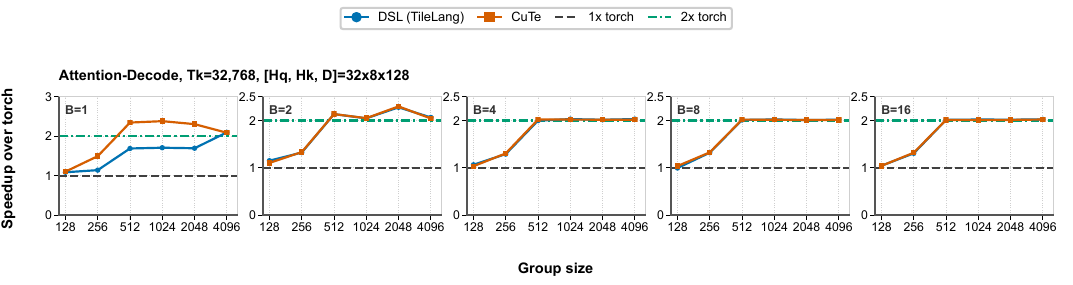}
    \caption{Attention decoding speedup for different group sizes on Llama3.1-8B shapes, with a random 50\% sparsity mask.}
    \label{fig:attention_speedup_vary_g}
\end{figure}

\paragraph{Decoding Speedup for Different Attention Group Sizes}
Figure~\ref{fig:attention_speedup_vary_g} shows how the attention group size affects decoding. For $G_{\text{attn}}\leqslant \frac{H_q}{H_k}\times d$, the kernel progressively transitions from vanilla multi-head attention to GQA-packing attention, where a single CTA partition can read and compute at most $\frac{H_q}{H_k}$ heads in parallel. Increasing $G_{\text{attn}}$ beyond this threshold does not affect the tiling strategy used for attention decoding.

\begin{figure}[htbp]
    \centering
    \includegraphics[width=0.9\linewidth]{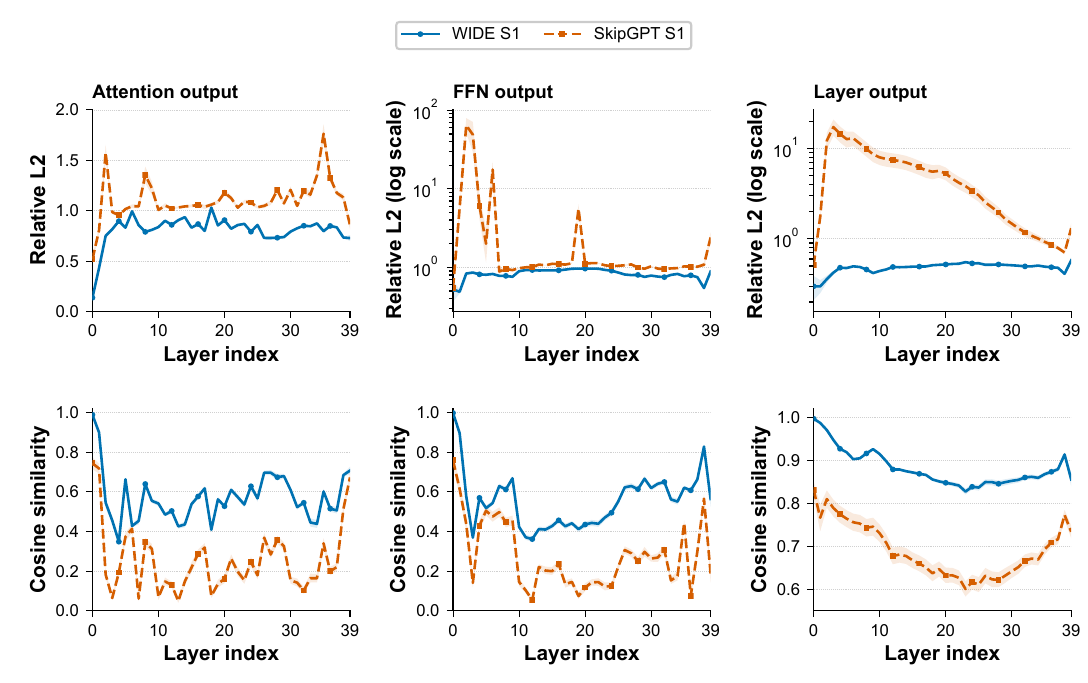}
    \caption{Layer-wise error accumulation in SkipGPT and \med~on Qwen3-14B after router training, measured by relative $\ell_2$ error and cosine similarity against the dense model.}
    \label{fig:layer_wise_error_s1}
\end{figure}

\begin{figure}[htbp]
    \centering
    \includegraphics[width=0.9\linewidth]{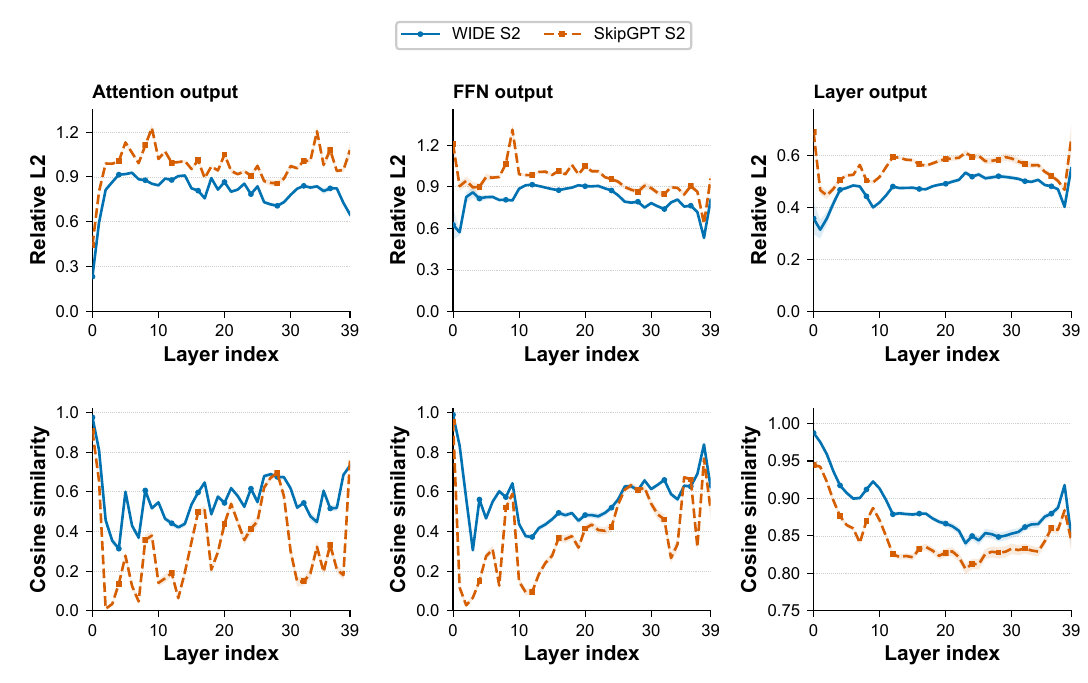}
    \caption{Layer-wise error accumulation in SkipGPT and \med~on Qwen3-14B after LoRA tuning.}
    \label{fig:layer_wise_error_s2}
\end{figure}

\paragraph{Analysis of Layer-wise Error Accumulation}
Beyond downstream task performance, we also examine how errors accumulate across layers in SkipGPT and \med~at 50\% sparsity compared to the dense baseline. Figures~\ref{fig:layer_wise_error_s1} and~\ref{fig:layer_wise_error_s2} report errors and similarity in the attention and FFN outputs, together with the accumulated error across each layer, with relative $\ell_2$ error defined as $\frac{\lVert\mathbf{X}_{\text{dense}} - \mathbf{X}_{\text{pruned}}\rVert_\text{F}}{\lVert\mathbf{X}_{\text{dense}}\rVert_\text{F}}$, where $\mathbf{X}$ denotes a layer's output tensor, and the subscripts denote the dense and pruned models. All evaluation uses 32 randomly sampled reference answers from GSM8K. It is evident that \med~consistently yields lower layer-wise accumulated errors than the coarser-grained SkipGPT across the two training stages.

\FloatBarrier

\begin{wrapfigure}{r}{0.45\linewidth}
    \vspace{0px}
    \centering
    \includegraphics[width=1.0\linewidth, trim=0 0 0 0, clip]{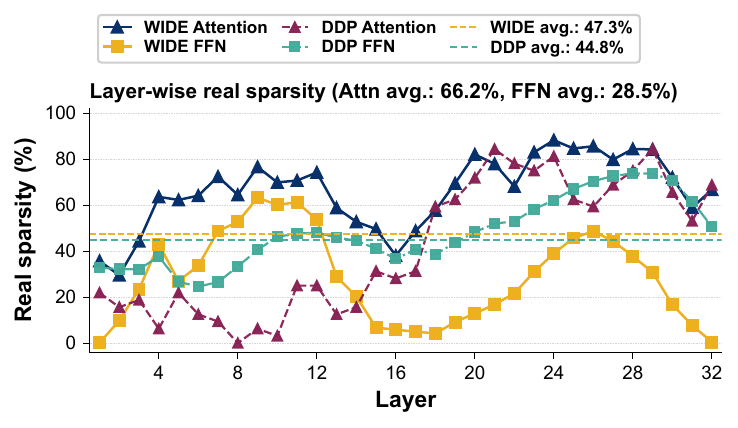}
    \caption{\med~and DDP's layer-wise sparsity distribution of Llama3.1-8B on WikiText2, with target sparsity set to 50\%.}
    \label{fig:layerwise_sparsity_llama31_wikitext}
\end{wrapfigure}
\begin{figure}[htbp]
    \centering
    \includegraphics[width=1.0\linewidth]{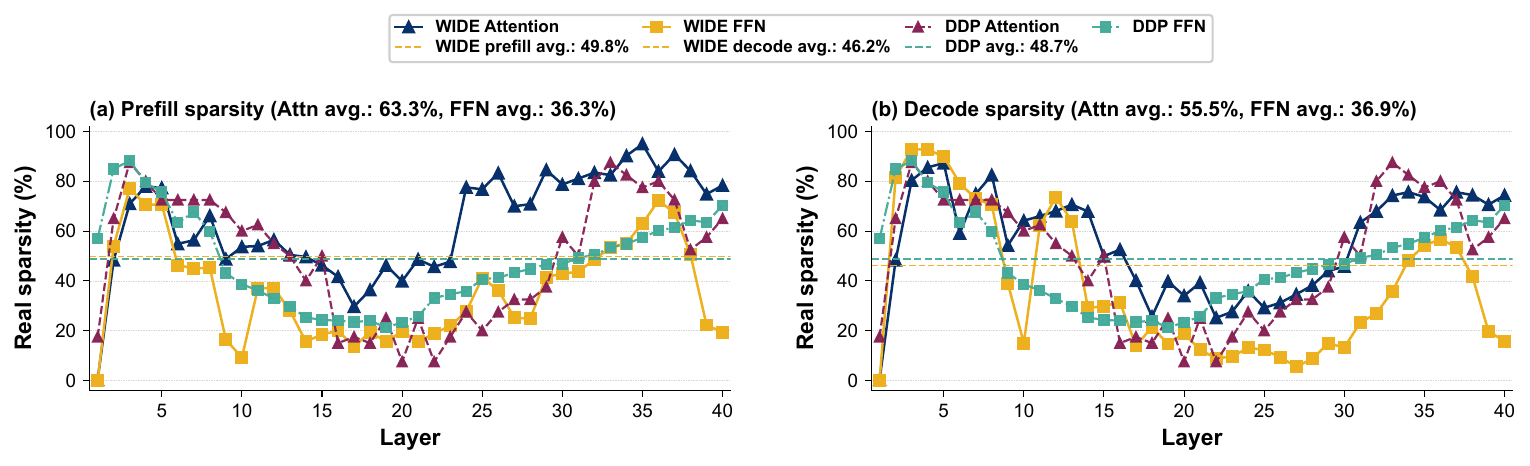}
    \caption{\med~and DDP's layer-wise sparsity distribution of Qwen3-14B on GSM8K's prefill and decoding stages, with target sparsity set to 50\%.}
    \label{fig:qwen3_gsm8k_prefill_decode_sparsity}
\end{figure}

\paragraph{Routing Behavior and Visualizations}
To understand how \med~allocates its compute budget across models and sparsity levels, we inspect the learned routing patterns at both the layer and token levels. All analyses use $G_{\text{ffn}}=128$, while the attention-side group size is aligned with the KV group of each model. For the layer-wise allocation, we compare \med~with the layer-wise strategy learned by DDP, using LoRA-tuned checkpoints for both methods. As shown in Fig.~\ref{fig:layerwise_sparsity_llama31_wikitext}, the learned allocation is highly non-uniform across modules and layers. Attention carries most of the skipping, reaching 66.2\% sparsity on WikiText2, while the FFN remains more conservative at 28.5\%. Across layers, both curves peak around layers 4--12 and 24--28, and dip near the model boundaries and layers 15--18. These profiles echo the layer sensitivity observed in static pruning studies~\citep{he2024AttnDrop,huang2026DDP}, but are produced here by token-conditioned group routing. The GSM8K analysis of Qwen3-14B further shows that \med~adopts model-dependent allocation patterns. As shown in Fig.~\ref{fig:qwen3_gsm8k_prefill_decode_sparsity}, Qwen3-14B instead applies higher FFN sparsity in layers 1--6, with the allocation also shifting between prefill and decoding.

\begin{figure}[htbp]
    \centering
    \includegraphics[width=1.0\linewidth]{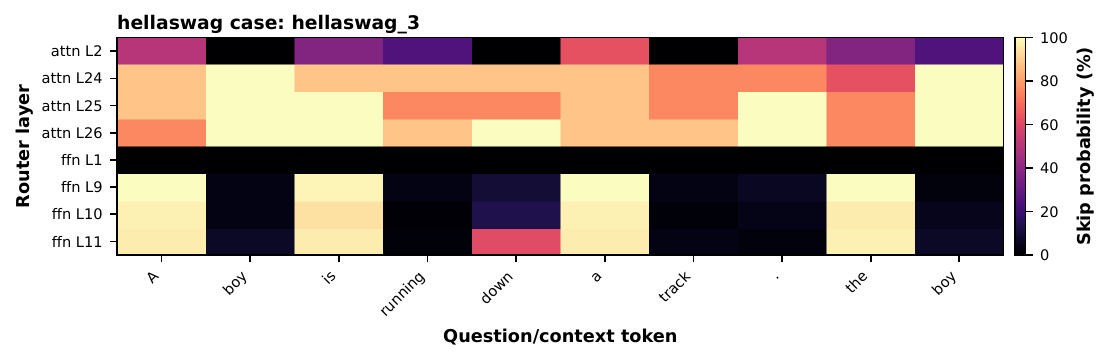}
    \caption{Token-wise routing probability for Llama3.1-8B in a selected HellaSwag sample.}
    \label{fig:semantic_viz_1}
\end{figure}

\begin{figure}[htbp]
    \centering
    \includegraphics[width=1.0\linewidth]{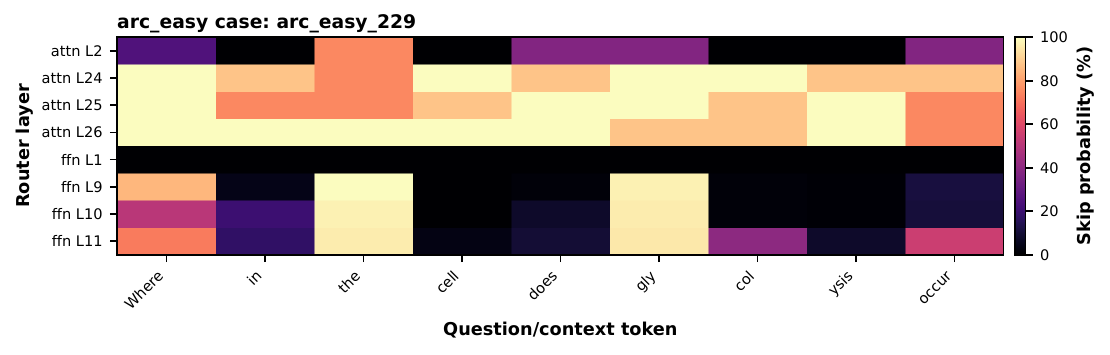}
    \caption{Token-wise routing probability for Llama3.1-8B in a selected ARC-Easy sample.}
    \label{fig:semantic_viz_2}
\end{figure}

\begin{figure}[htbp]
    \centering
    \includegraphics[width=1.0\linewidth]{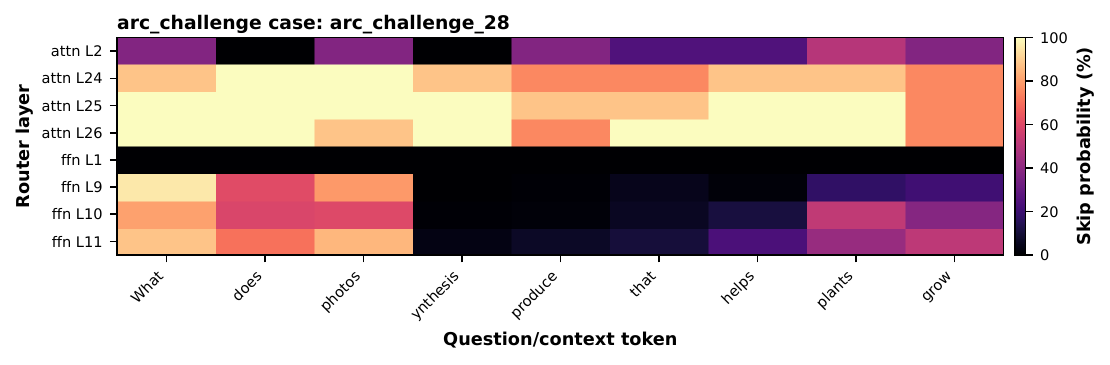}
    \caption{Token-wise routing probability for Llama3.1-8B in a selected ARC-Challenge sample.}
    \label{fig:semantic_viz_3}
\end{figure}

\begin{figure}[htbp]
    \centering
    \includegraphics[width=1.0\linewidth]{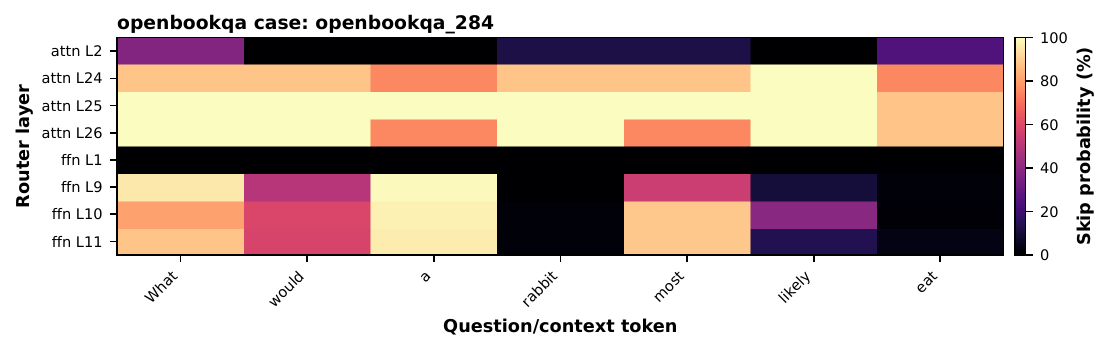}
    \caption{Token-wise routing probability for Llama3.1-8B in a selected OpenBookQA sample.}
    \label{fig:semantic_viz_4}
\end{figure}

\begin{figure}[htbp]
    \centering
    \includegraphics[width=1.0\linewidth]{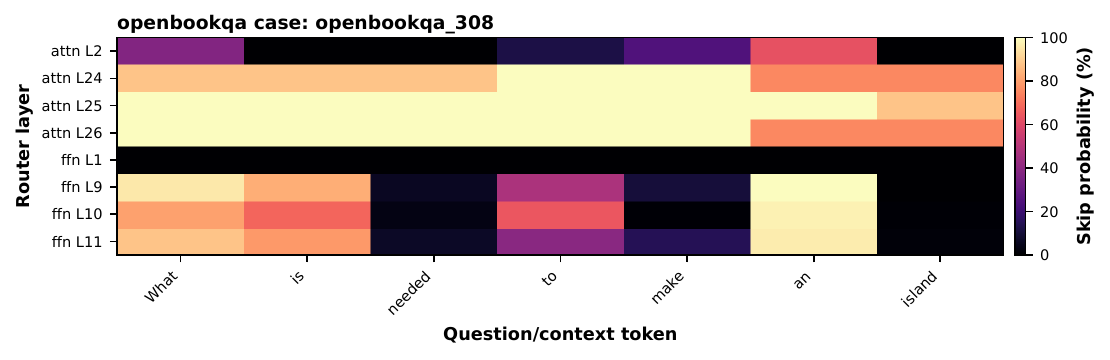}
    \caption{Token-wise routing probability for Llama3.1-8B in a selected OpenBookQA sample.}
    \label{fig:semantic_viz_5}
\end{figure}

\begin{figure}[htbp]
    \centering
    \includegraphics[width=1.0\linewidth]{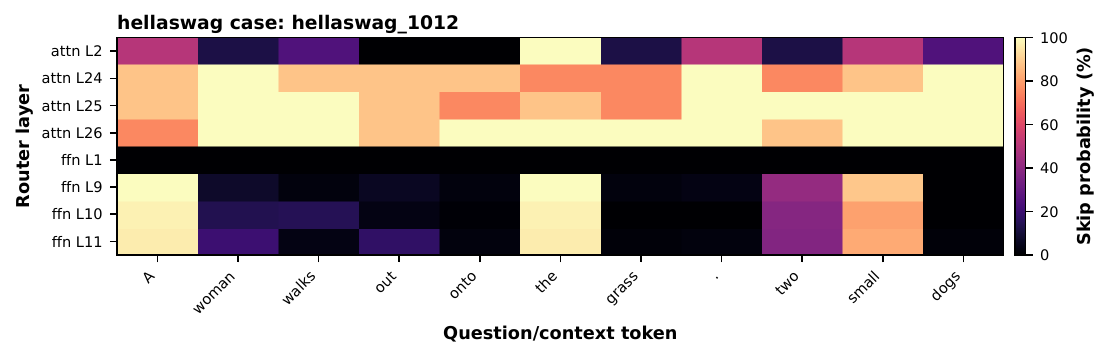}
    \caption{Token-wise routing probability for Llama3.1-8B in a selected HellaSwag sample.}
    \label{fig:semantic_viz_6}
\end{figure}

For the token-wise routing behavior, Figures~\ref{fig:semantic_viz_1} to~\ref{fig:semantic_viz_6} show several representative case studies on evaluation tasks. Here, the FFN routers keep semantic content tokens such as \texttt{boy} and \texttt{track}, especially in layers 9--11 where the average FFN sparsity is high, while function words like \texttt{that} and \texttt{the} are often skipped. This analysis highlights the ability of \med~to dynamically allocate its budget across layers and tokens based on semantic content.

\begin{figure}[htbp]
    \centering
    \includegraphics[width=1.0\linewidth]{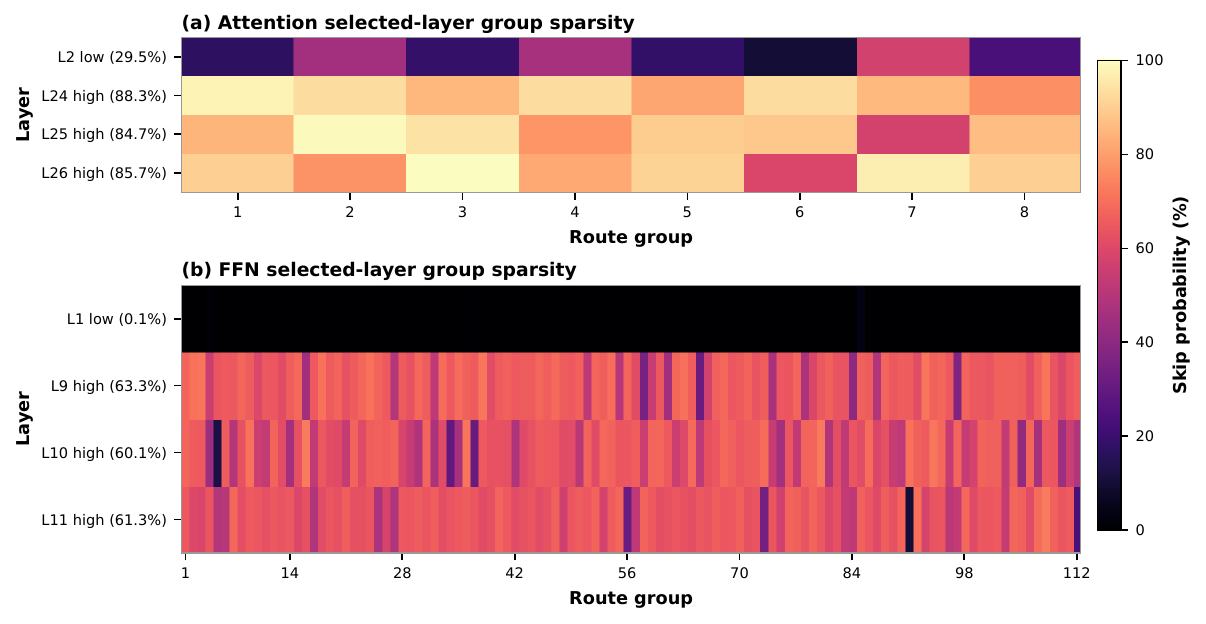}
    \caption{Group-wise skipping probabilities in different layers of Llama3.1-8B on WikiText2.}
    \label{fig:group_viz}
\end{figure}

As for the layer-wise group routing preferences, Figure~\ref{fig:group_viz} further compares the probability of skipping each routing group across low and high sparsity layers in the attention and FFN modules, where the patterns differ across layers.

\FloatBarrier

\section{Full Downstream Benchmark Results}
Table~\ref{tbl:full_bench_llama31-8b_25},~\ref{tbl:full_bench_llama31-8b_50}, ~\ref{tbl:full_bench_llama32-3b_50}, ~\ref{tbl:full_bench_qwen3-14b_25}, and~\ref{tbl:full_bench_qwen3-14b_50} report the full results for Llama3.1-8B, Llama3.2-3B, and Qwen3-14B at the task level.
\begin{table}[htbp]
    \centering
    \caption{Llama3.1-8B performance with target sparsity 25\%.}
    \aboverulesep=0ex
    \belowrulesep=0ex
    \renewcommand{\arraystretch}{1.2}
    \resizebox{1.0\linewidth}{!}{
        \begin{tabular}{l |c|ccccccc|c}
            \toprule
            \multirow{2}{*}{Methods} & WikiText2 & ARC-E & ARC-C & BoolQ & Winogrande & PIQA & OpenbookQA & Hellaswag & \multirow{2}{*}{Avg. Acc.$\uparrow$}\\
            & (ppl$\downarrow$) & (Acc. Norm.$\uparrow$) & (Acc. Norm.$\uparrow$) & (Acc. $\uparrow$) & (Acc. $\uparrow$) & (Acc. Norm.$\uparrow$) & (Acc. Norm.$\uparrow$) & (Acc. Norm.$\uparrow$) &\\
            \midrule
            \multicolumn{10}{c}{\textbf{Calibration-only (S1)}}\\
            \midrule
            Dense & 7.71 & 82.70 & 55.03 & 83.06 & 74.27 & 81.07 & 45.40 & 79.32 & 71.55 (100.00\%) \\
            \midrule
            \rowcolor{gray!15}
            Shortened-ppl & 25.84 & 52.19 & 29.95 & 41.41 & 54.62 & 69.80 & 33.00 & 52.55 & 47.65 (66.59\%) \\
            \rowcolor{gray!15}
            Shortened-taylor & 30.07 & 54.42 & 32.42 & 40.28 & 55.49 & 70.62 & 35.20 & 56.68 & 49.30 (68.90\%) \\
            \rowcolor{gray!15}
            CoopPruner & 22.96 & 59.81 & 33.79 & 61.56 & 53.99 & 72.52 & 33.60 & 58.06 & 53.33 (74.54\%) \\
            \rowcolor{orange!15}
            SliceGPT & 22.57 & 46.42 & 29.52 & 74.92 & 67.48 & 66.53 & 31.80 & 56.93 & 53.37 (74.59\%) \\
            \rowcolor{orange!15}
            T\'yr-the-Pruner & 12.89 & 68.22 & 38.73 & 78.59 & 66.53 & 74.21 & 37.40 & 66.33 & 61.43 (85.86\%) \\
            \rowcolor{orange!15}
            DDP & 12.41 & 70.01 & 43.26 & 77.25 & \underline{69.85} & 75.63 & \underline{42.00} & 72.42 & 64.35 (89.93\%) \\
            \rowcolor{cyan!15}
            D-LLM & 69.45 & 29.63 & 22.53 & 58.78 & 49.96 & 54.19 & 25.80 & 31.32 & 38.89 (54.35\%) \\
            \rowcolor{cyan!15}
            SkipGPT & 15.63 & 50.38 & 30.89 & 69.94 & 51.30 & 68.77 & 31.40 & 59.06 & 51.68 (72.23\%) \\
            \rowcolor{cyan!15}
            \med~(32) & \textbf{8.49} & \textbf{79.84} & \textbf{52.82} & \textbf{82.17} & \textbf{71.67} & \textbf{79.76} & \textbf{45.60} & \textbf{78.58} & \textbf{70.06 (97.92\%)} \\
            \rowcolor{cyan!15}
            \med~(64) & 10.37 & \underline{71.34} & \underline{44.97} & \underline{80.46} & 62.15 & 79.11 & 40.40 & \underline{75.78} & \underline{64.89 (90.69\%)} \\
            \rowcolor{cyan!15}
            \med~(128) & \underline{10.04} & 68.77 & \underline{44.97} & 77.95 & 60.77 & \underline{79.16} & 38.20 & 75.73 & 63.65 (88.96\%) \\
            \midrule
            \multicolumn{10}{c}{\textbf{LoRA Tuning (S2)}}\\
            \midrule
            Dense & 7.91 & 81.52 & 54.01 & 83.30 & 73.32 & 81.23 & 46.20 & 79.94 & 71.36 (99.73\%) \\
            \midrule
            \rowcolor{gray!15}
            Shortened-ppl & 12.33 & 67.63 & 39.08 & 54.25 & 58.48 & 76.61 & 39.00 & 65.69 & 57.25 (80.01\%) \\
            \rowcolor{gray!15}
            Shortened-taylor & 12.12 & 70.58 & 40.78 & 64.04 & 59.83 & 76.88 & 38.20 & 68.53 & 59.83 (83.63\%) \\
            \rowcolor{gray!15}
            CoopPruner & 12.03 & 73.53 & 42.41 & 63.52 & 62.51 & 77.04 & 40.00 & 69.76 & 61.25 (85.61\%) \\
            \rowcolor{orange!15}
            SliceGPT & 14.88 & 62.92 & 36.86 & 76.17 & 68.82 & 73.77 & 36.80 & 69.71 & 60.72 (84.87\%) \\
            \rowcolor{orange!15}
            T\'yr-the-Pruner & 11.49 & 71.25 & 43.17 & 78.50 & 68.74 & 76.27 & 39.80 & 71.68 & 64.20 (89.73\%) \\
            \rowcolor{orange!15}
            DDP & 12.10 & 70.16 & 43.69 & 78.78 & 68.35 & 76.01 & 40.60 & 72.12 & 64.24 (89.79\%) \\
            \rowcolor{cyan!15}
            D-LLM & 10.29 & 77.36 & 48.38 & 76.39 & 68.27 & 78.29 & 43.80 & 75.38 & 66.84 (93.42\%) \\
            \rowcolor{cyan!15}
            SkipGPT & \underline{9.87} & \textbf{80.68} & 51.96 & \underline{81.47} & \underline{70.88} & 79.65 & 44.00 & 77.73 & 69.48 (97.11\%) \\
            \rowcolor{cyan!15}
            \med~(32) & \textbf{8.61} & \underline{80.30} & \textbf{53.50} & \textbf{82.66} & \underline{70.88} & 80.30 & \underline{45.20} & \textbf{78.46} & \textbf{70.19 (98.09\%)} \\
            \rowcolor{cyan!15}
            \med~(64) & 10.32 & 79.50 & \underline{52.56} & \underline{81.47} & \textbf{71.19} & \textbf{80.52} & \underline{45.20} & 78.34 & \underline{69.83 (97.59\%)} \\
            \rowcolor{cyan!15}
            \med~(128) & 12.51 & 79.29 & 51.45 & 80.58 & \underline{70.88} & \underline{80.36} & \textbf{45.80} & \underline{78.35} & 69.53 (97.18\%) \\
            \bottomrule
        \end{tabular}
    }
    \label{tbl:full_bench_llama31-8b_25}
\end{table}

\begin{table}[htbp]
    \centering
    \caption{Llama3.1-8B performance with target sparsity 50\%.}
    \aboverulesep=0ex
    \belowrulesep=0ex
    \renewcommand{\arraystretch}{1.2}
    \resizebox{1.0\linewidth}{!}{
        \begin{tabular}{l |c|ccccccc|c}
            \toprule
            \multirow{2}{*}{Methods} & WikiText2 & ARC-E & ARC-C & BoolQ & Winogrande & PIQA & OpenbookQA & Hellaswag & \multirow{2}{*}{Avg. Acc.$\uparrow$}\\
            & (ppl$\downarrow$) & (Acc. Norm.$\uparrow$) & (Acc. Norm.$\uparrow$) & (Acc. $\uparrow$) & (Acc. $\uparrow$) & (Acc. Norm.$\uparrow$) & (Acc. Norm.$\uparrow$) & (Acc. Norm.$\uparrow$) &\\
            \midrule
            \multicolumn{10}{c}{\textbf{Calibration-only (S1)}}\\
            \midrule
            Dense & 7.71 & 82.70 & 55.03 & 83.06 & 74.27 & 81.07 & 45.40 & 79.32 & 71.55 (100.00\%) \\
            \midrule
            \rowcolor{gray!15}
            Shortened-ppl & 473.15 & 32.20 & 24.40 & 53.00 & 49.49 & 57.40 & 25.80 & 35.03 & 39.62 (55.37\%) \\
            \rowcolor{gray!15}
            Shortened-taylor & 4.86e+7 & 29.17 & 25.94 & 38.20 & 49.88 & 53.70 & 27.40 & 27.63 & 35.99 (50.30\%) \\
            \rowcolor{gray!15}
            CoopPruner & 503.86 & 32.87 & 23.21 & 56.09 & 49.33 & 56.80 & 26.20 & 30.45 & 39.28 (54.90\%) \\
            \rowcolor{orange!15}
            SliceGPT & 85.81 & 30.30 & 21.84 & 55.47 & 51.53 & 53.97 & 25.80 & 32.05 & 38.71 (54.10\%) \\
            \rowcolor{orange!15}
            T\'yr-the-Pruner & 59.88 & 44.65 & 24.91 & 58.44 & 54.61 & 64.68 & 29.40 & 41.62 & 45.47 (63.55\%) \\
            \rowcolor{orange!15}
            DDP & 21.83 & 55.13 & 33.36 & 65.69 & 60.77 & 69.31 & 30.80 & 56.28 & 53.05 (74.14\%) \\
            \rowcolor{cyan!15}
            D-LLM & 504.48 & 27.02 & 23.12 & 50.52 & 47.20 & 52.72 & 25.20 & 28.48 & 36.32 (50.77\%) \\
            \rowcolor{cyan!15}
            SkipGPT & 96.04 & 35.69 & 24.57 & 57.65 & 52.49 & 58.71 & 26.60 & 41.92 & 42.52 (59.42\%) \\
            \rowcolor{cyan!15}
            \med~(32) & \underline{14.15} & \textbf{69.74} & \textbf{41.55} & \textbf{70.31} & 63.69 & \underline{76.22} & \textbf{42.20} & \textbf{69.17} & \textbf{61.84 (86.43\%)} \\
            \rowcolor{cyan!15}
            \med~(64) & 14.96 & \underline{69.28} & 38.99 & 67.43 & \underline{66.69} & 75.95 & \underline{41.40} & 68.50 & 61.18 (85.50\%) \\
            \rowcolor{cyan!15}
            \med~(128) & \textbf{12.57} & 66.37 & \underline{39.33} & \underline{69.51} & \textbf{67.32} & \textbf{76.55} & \textbf{42.20} & \underline{69.14} & \underline{61.49 (85.94\%)} \\
            \midrule
            \multicolumn{10}{c}{\textbf{LoRA Tuning (S2)}}\\
            \midrule
            Dense & 7.91 & 81.52 & 54.01 & 83.30 & 73.32 & 81.23 & 46.20 & 79.94 & 71.36 (99.73\%) \\
            \midrule
            \rowcolor{gray!15}
            Shortened-ppl & 22.47 & 51.56 & 27.56 & 50.37 & 53.75 & 68.12 & 32.00 & 47.44 & 47.26 (66.05\%) \\
            \rowcolor{gray!15}
            Shortened-taylor & 22.30 & 52.57 & 29.69 & 61.01 & 54.62 & 69.75 & 33.20 & 48.92 & 49.97 (69.83\%) \\
            \rowcolor{gray!15}
            CoopPruner & 22.66 & 52.82 & 29.18 & 63.15 & 54.30 & 67.95 & 30.20 & 49.80 & 49.63 (69.36\%) \\
            \rowcolor{orange!15}
            SliceGPT & 28.76 & 39.64 & 24.06 & 64.12 & 58.24 & 60.60 & 28.00 & 47.43 & 46.01 (64.31\%) \\
            \rowcolor{orange!15}
            T\'yr-the-Pruner & 37.99 & 54.75 & 31.74 & 68.56 & 58.72 & 70.56 & 33.80 & 56.09 & 53.46 (74.72\%) \\
            \rowcolor{orange!15}
            DDP & 19.52 & 55.77 & 34.56 & 68.26 & 61.80 & 71.00 & 33.60 & 57.46 & 54.64 (76.36\%) \\
            \rowcolor{cyan!15}
            D-LLM & 19.31 & 63.05 & 35.75 & 62.45 & 60.14 & 72.36 & 37.40 & 61.31 & 56.07 (78.36\%) \\
            \rowcolor{cyan!15}
            SkipGPT & 13.90 & 72.01 & \underline{42.24} & 65.81 & 64.80 & 75.68 & 42.80 & 67.87 & 61.60 (86.10\%) \\
            \rowcolor{cyan!15}
            \med~(32) & \underline{11.99} & \underline{72.90} & \textbf{44.20} & \underline{77.22} & 65.98 & \underline{78.40} & \underline{43.40} & 71.66 & \textbf{64.82 (90.60\%)} \\
            \rowcolor{cyan!15}
            \med~(64) & 12.05 & \textbf{73.91} & 41.72 & \textbf{77.74} & \underline{66.30} & 78.07 & 42.80 & \textbf{72.03} & \underline{64.65 (90.36\%)} \\
            \rowcolor{cyan!15}
            \med~(128) & \textbf{11.43} & 71.04 & 41.64 & 75.38 & \textbf{67.72} & \textbf{78.94} & \textbf{43.80} & \underline{71.85} & 64.34 (89.92\%) \\
            \bottomrule
        \end{tabular}
    }
    \label{tbl:full_bench_llama31-8b_50}
\end{table}

\begin{table}[htbp]
    \centering
    \caption{Llama3.2-3B performance with target sparsity 50\%.}
    \aboverulesep=0ex
    \belowrulesep=0ex
    \renewcommand{\arraystretch}{1.2}
    \resizebox{1.0\linewidth}{!}{
        \begin{tabular}{l |c|ccccccc|c}
            \toprule
            \multirow{2}{*}{Methods} & WikiText2 & ARC-E & ARC-C & BoolQ & Winogrande & PIQA & OpenbookQA & Hellaswag & \multirow{2}{*}{Avg. Acc.$\uparrow$}\\
            & (ppl$\downarrow$) & (Acc. Norm.$\uparrow$) & (Acc. Norm.$\uparrow$) & (Acc. $\uparrow$) & (Acc. $\uparrow$) & (Acc. Norm.$\uparrow$) & (Acc. Norm.$\uparrow$) & (Acc. Norm.$\uparrow$) &\\
            \midrule
            \multicolumn{10}{c}{\textbf{Calibration-only (S1)}}\\
            \midrule
            Dense & 9.77 & 72.01 & 46.67 & 73.88 & 69.22 & 78.02 & 40.60 & 73.98 & 64.91 (100.00\%) \\
            \midrule
            \rowcolor{gray!15}
            Shortened-ppl & 676.05 & 33.88 & 23.55 & 44.19 & 52.75 & 56.58 & 25.20 & 31.44 & 38.23 (58.89\%) \\
            \rowcolor{gray!15}
            Shortened-taylor & 2.32e+5 & 34.01 & 25.60 & 50.92 & 51.93 & 55.71 & 25.60 & 31.70 & 39.35 (60.63\%) \\
            \rowcolor{gray!15}
            CoopPruner & 1.79e+5 & 25.55 & 24.57 & 50.67 & 50.04 & 52.23 & 27.00 & 26.25 & 36.62 (56.41\%) \\
            \rowcolor{orange!15}
            SliceGPT & 96.69 & 29.12 & 21.67 & 54.58 & 48.61 & 53.21 & 25.80 & 30.38 & 37.62 (57.96\%) \\
            \rowcolor{orange!15}
            T\'yr-the-Pruner & 87.37 & 38.38 & 23.20 & 62.17 & 53.35 & 59.63 & 27.40 & 34.06 & 42.60 (65.63\%) \\
            \rowcolor{orange!15}
            DDP & 31.20 & 47.14 & 29.78 & 62.42 & 56.51 & 65.34 & 30.00 & 45.04 & 48.03 (74.00\%) \\
            \rowcolor{cyan!15}
            D-LLM & 3877.53 & 27.19 & 24.32 & 40.86 & 49.57 & 50.65 & 26.00 & 26.35 & 34.99 (53.91\%) \\
            \rowcolor{cyan!15}
            SkipGPT & 156.58 & 30.89 & 20.99 & 43.30 & 50.04 & 55.77 & 25.80 & 30.38 & 36.74 (56.60\%) \\
            \rowcolor{cyan!15}
            \med~(32) & \textbf{17.28} & \textbf{62.84} & \textbf{35.41} & \underline{65.20} & \textbf{62.51} & \underline{73.01} & 38.40 & \textbf{61.60} & \textbf{57.00 (87.81\%)} \\
            \rowcolor{cyan!15}
            \med~(64) & 19.02 & \underline{61.74} & 34.64 & \textbf{66.67} & 59.98 & 72.36 & \underline{38.60} & \underline{60.84} & \underline{56.40 (86.89\%)} \\
            \rowcolor{cyan!15}
            \med~(128) & \underline{18.11} & 57.66 & \underline{35.32} & 62.02 & \underline{60.77} & \textbf{73.23} & \textbf{39.20} & 59.76 & 55.42 (85.38\%) \\
            \midrule
            \multicolumn{10}{c}{\textbf{LoRA Tuning (S2)}}\\
            \midrule
            Dense & 9.86 & 72.64 & 46.42 & 72.87 & 68.98 & 77.91 & 42.20 & 74.00 & 65.00 (100.14\%) \\
            \midrule
            \rowcolor{gray!15}
            Shortened-ppl & 29.75 & 46.51 & 27.47 & 61.28 & 51.22 & 64.85 & 29.60 & 39.91 & 45.83 (70.61\%) \\
            \rowcolor{gray!15}
            Shortened-taylor & 31.50 & 46.13 & 25.94 & 61.88 & 51.70 & 64.85 & 28.20 & 42.16 & 45.84 (70.61\%) \\
            \rowcolor{gray!15}
            CoopPruner & 42.69 & 39.23 & 23.12 & 49.24 & 52.96 & 63.11 & 29.20 & 35.89 & 41.82 (64.43\%) \\
            \rowcolor{orange!15}
            SliceGPT & 31.13 & 36.48 & 22.44 & 59.32 & 54.61 & 59.46 & 26.60 & 40.40 & 42.76 (65.87\%) \\
            \rowcolor{orange!15}
            T\'yr-the-Pruner & 53.11 & 45.62 & 25.25 & 57.58 & 52.56 & 65.34 & 30.00 & 43.46 & 45.69 (70.38\%) \\
            \rowcolor{orange!15}
            DDP & 25.66 & 48.36 & 30.20 & 63.30 & 56.12 & 67.14 & 30.20 & 46.84 & 48.88 (75.30\%) \\
            \rowcolor{cyan!15}
            D-LLM & 193.97 & 31.40 & 21.25 & 57.09 & 51.22 & 57.18 & 24.40 & 28.07 & 38.66 (59.56\%) \\
            \rowcolor{cyan!15}
            SkipGPT & 18.91 & 58.92 & 33.45 & 62.81 & 58.64 & 72.31 & 36.60 & 57.59 & 54.33 (83.70\%) \\
            \rowcolor{cyan!15}
            \med~(32) & 15.28 & \textbf{67.72} & \textbf{37.97} & \textbf{66.94} & 59.27 & 74.54 & \textbf{40.60} & \underline{64.32} & \textbf{58.77 (90.53\%)} \\
            \rowcolor{cyan!15}
            \med~(64) & \textbf{15.17} & \underline{65.78} & \underline{37.29} & 65.23 & \underline{61.01} & \textbf{75.14} & \underline{39.80} & 64.26 & \underline{58.36 (89.90\%)} \\
            \rowcolor{cyan!15}
            \med~(128) & \underline{15.26} & 65.74 & 36.43 & \underline{66.12} & \textbf{61.88} & \underline{74.92} & 37.00 & \textbf{64.69} & 58.11 (89.52\%) \\
            \bottomrule
        \end{tabular}
    }
    \label{tbl:full_bench_llama32-3b_50}
\end{table}

\begin{table}[htbp]
    \centering
    \caption{Qwen3-14B performance with target sparsity 25\%.}
    \aboverulesep=0ex
    \belowrulesep=0ex
    \renewcommand{\arraystretch}{1.2}
    \resizebox{1.0\linewidth}{!}{
        \begin{tabular}{l|c|cccccccc|ccc|cc}
            \toprule
            \multirow{2}{*}{Methods} & WikiText2 & ARC-E & ARC-C & BoolQ & Winogrande & PIQA & OpenbookQA & Hellaswag & MMLU & IFEval & MBPP+ & GSM8K & \multirow{2}{*}{Like. Acc.} & \multirow{2}{*}{Gen. Acc.}\\
            & & & & & & & & & (5-shot Acc.$\uparrow$) & (prompt-strict $\uparrow$) & (pass@1$\uparrow$) & (Acc.$\uparrow$) & &\\
            \midrule
            \multicolumn{15}{c}{\textbf{Calibration-only (S1)}}\\
            \midrule
            Dense & 11.12 & 82.79 & 59.98 & 89.33 & 72.69 & 79.82 & 46.40 & 78.78 & 78.79 & 85.39 & 89.68 & 79.30 & 73.57 (100.00\%) & 84.79 (100.00\%) \\
            \midrule
            \rowcolor{gray!15}
            Shortened-ppl & 30.71 & 43.56 & 28.67 & 61.80 & 50.99 & 67.46 & 33.40 & 47.35 & 23.19 & 10.35 & 1.06 & 1.59 & 44.55 (60.55\%) & 4.33 (5.11\%) \\
            \rowcolor{gray!15}
            Shortened-taylor & 9970.21 & 28.37 & 23.63 & 60.58 & 50.36 & 51.47 & 30.60 & 25.91 & 23.20 & 14.79 & 0.00 & 0.00 & 36.76 (49.97\%) & 4.93 (5.81\%) \\
            \rowcolor{gray!15}
            CoopPruner & 29.32 & 46.30 & 29.95 & 57.89 & 52.09 & 66.70 & 33.20 & 44.81 & 22.95 & 13.12 & 1.32 & 3.34 & 44.23 (60.12\%) & 5.92 (6.98\%) \\
            \rowcolor{orange!15}
            DDP & 16.87 & 73.95 & 52.47 & 88.01 & 66.06 & 75.03 & 40.40 & 63.59 & 62.64 & \textbf{70.79} & 80.69 & 68.23 & 65.26 (88.71\%) & 73.23 (86.37\%) \\
            \rowcolor{cyan!15}
            SkipGPT & 20.86 & 75.92 & 52.81 & 84.03 & 58.48 & 75.29 & 37.80 & 63.96 & 44.79 & 48.24 & 1.05 & 65.73 & 61.63 (83.77\%) & 38.34 (45.21\%) \\
            \rowcolor{cyan!15}
            \med~(32) & \textbf{11.85} & 85.69 & \underline{63.39} & \underline{88.34} & \textbf{70.63} & 80.08 & 45.80 & 74.23 & \underline{73.50} & 69.87 & 81.74 & \underline{82.33} & 72.70 (98.82\%) & \underline{77.98 (91.96\%)} \\
            \rowcolor{cyan!15}
            \med~(64) & 15.63 & \textbf{86.82} & \textbf{66.21} & 87.40 & 68.66 & \textbf{80.63} & \textbf{48.20} & \underline{74.78} & 72.89 & \underline{70.60} & \underline{82.01} & 81.34 & \textbf{73.19 (99.49\%)} & \underline{77.98 (91.97\%)} \\
            \rowcolor{cyan!15}
            \med~(128) & \underline{12.15} & \underline{86.53} & 62.96 & \textbf{88.65} & \underline{69.61} & \underline{80.35} & \underline{48.00} & \textbf{74.88} & \textbf{73.94} & 70.42 & \textbf{82.53} & \textbf{82.78} & \underline{73.11 (99.37\%)} & \textbf{78.57 (92.67\%)} \\
            \midrule
            \multicolumn{15}{c}{\textbf{LoRA Tuning (S2)}}\\
            \midrule
            Dense & 9.58 & 87.84 & 68.09 & 88.99 & 74.59 & 82.21 & 48.60 & 80.32 & 78.69 & 78.18 & 84.65 & 88.77 & 76.16 (103.52\%) & 83.86 (98.91\%) \\
            \midrule
            \rowcolor{gray!15}
            Shortened-ppl & 14.65 & 77.57 & 49.83 & 72.48 & 60.38 & 77.20 & 43.00 & 62.73 & 36.08 & 29.57 & 46.83 & 45.26 & 59.90 (81.42\%) & 40.55 (47.82\%) \\
            \rowcolor{gray!15}
            Shortened-taylor & 88.33 & 75.59 & 52.90 & 85.35 & 66.22 & 73.23 & 40.20 & 65.78 & 62.83 & 58.78 & 69.05 & 73.46 & 65.26 (88.70\%) & 67.09 (79.13\%) \\
            \rowcolor{gray!15}
            CoopPruner & 14.21 & 78.54 & 51.54 & 77.71 & 64.80 & 76.71 & 42.00 & 64.86 & 46.97 & 43.81 & 53.97 & 58.83 & 62.89 (85.48\%) & 52.20 (61.56\%) \\
            \rowcolor{orange!15}
            DDP & 15.54 & 80.18 & 58.36 & 87.43 & 69.06 & 76.17 & 42.40 & 68.97 & 67.66 & 66.36 & 76.98 & 80.52 & 68.77 (93.48\%) & 74.62 (88.00\%) \\
            \rowcolor{cyan!15}
            SkipGPT & 12.30 & 86.19 & 63.56 & 87.64 & 64.95 & \textbf{81.17} & \underline{49.00} & 74.67 & 68.50 & 64.69 & 79.36 & 85.97 & 71.96 (97.80\%) & 76.67 (90.42\%) \\
            \rowcolor{cyan!15}
            \med~(32) & \textbf{11.10} & 86.86 & \underline{65.52} & 88.37 & \textbf{71.66} & 80.57 & 48.80 & 76.14 & 74.17 & \textbf{71.16} & \underline{83.59} & 87.94 & 74.01 (100.59\%) & \underline{80.89 (95.40\%)} \\
            \rowcolor{cyan!15}
            \med~(64) & 14.99 & \underline{87.20} & 65.27 & \textbf{89.11} & \underline{70.95} & \underline{80.90} & 49.00 & \underline{76.20} & \underline{75.13} & \underline{68.94} & \textbf{83.86} & \textbf{89.61} & \underline{74.22 (100.88\%)} & 80.80 (95.29\%) \\
            \rowcolor{cyan!15}
            \med~(128) & \underline{11.15} & \textbf{87.45} & \textbf{65.69} & \underline{89.08} & 69.29 & 80.79 & \textbf{50.40} & \textbf{76.73} & \textbf{75.24} & \textbf{71.16} & \textbf{83.86} & \underline{89.38} & \textbf{74.33 (101.03\%)} & \textbf{81.46 (96.08\%)} \\
            \bottomrule
        \end{tabular}
    }
    \label{tbl:full_bench_qwen3-14b_25}
\end{table}

\begin{table}[htbp]
    \centering
    \caption{Qwen3-14B performance with target sparsity 50\%.}
    \aboverulesep=0ex
    \belowrulesep=0ex
    \renewcommand{\arraystretch}{1.2}
    \resizebox{1.0\linewidth}{!}{
        \begin{tabular}{l|c|cccccccc|ccc|cc}
            \toprule
            \multirow{2}{*}{Methods} & WikiText2 & ARC-E & ARC-C & BoolQ & Winogrande & PIQA & OpenbookQA & Hellaswag & MMLU & IFEval & MBPP+ & GSM8K & \multirow{2}{*}{Like. Acc.} & \multirow{2}{*}{Gen. Acc.}\\
            & & & & & & & & & (5-shot Acc.$\uparrow$) & (prompt-strict $\uparrow$) & (pass@1$\uparrow$) & (Acc.$\uparrow$) & &\\
            \midrule
            \multicolumn{15}{c}{\textbf{Calibration-only (S1)}}\\
            \midrule
            Dense & 11.12 & 82.79 & 59.98 & 89.33 & 72.69 & 79.82 & 46.40 & 78.78 & 78.79 & 85.39 & 89.68 & 79.30 & 73.57 (100.00\%) & 84.79 (100.00\%) \\
            \midrule
            \rowcolor{gray!15}
            Shortened-ppl & 5.41e+5 & 26.38 & 26.36 & 42.23 & 49.96 & 51.25 & 29.80 & 25.26 & 23.91 & 8.68 & 0.00 & 0.98 & 34.39 (46.74\%) & 3.22 (3.79\%) \\
            \rowcolor{gray!15}
            Shortened-taylor & 2.27e+7 & 27.23 & 26.53 & 47.46 & 49.25 & 51.08 & 30.60 & 25.54 & 26.38 & 11.09 & 0.00 & 0.30 & 35.50 (48.26\%) & 3.79 (4.47\%) \\
            \rowcolor{gray!15}
            CoopPruner & 21058.12 & 28.91 & 24.82 & 40.88 & 51.30 & 53.42 & 32.40 & 26.93 & 26.63 & 13.30 & 0.00 & 0.07 & 35.66 (48.47\%) & 4.45 (5.25\%) \\
            \rowcolor{orange!15}
            DDP & 48.39 & 55.55 & 36.51 & 81.59 & 57.06 & 66.70 & 34.40 & 49.61 & 42.60 & 46.76 & 50.79 & 59.05 & 53.00 (72.04\%) & 52.20 (61.56\%) \\
            \rowcolor{cyan!15}
            SkipGPT & 546.29 & 32.19 & 26.45 & 39.02 & 51.14 & 51.79 & 27.20 & 28.54 & 24.87 & 23.65 & 0.26 & 0.37 & 35.15 (47.77\%) & 8.09 (9.54\%) \\
            \rowcolor{cyan!15}
            \med~(32) & \textbf{19.18} & 83.62 & 60.15 & \textbf{87.58} & \underline{65.82} & 76.27 & \underline{46.40} & 67.50 & 64.27 & \textbf{58.41} & \textbf{76.98} & \underline{79.07} & 68.95 (93.71\%) & \textbf{71.48 (84.31\%)} \\
            \rowcolor{cyan!15}
            \med~(64) & \underline{19.79} & \textbf{85.10} & \textbf{63.13} & 87.43 & \textbf{67.08} & \textbf{78.29} & \textbf{47.20} & \underline{68.11} & \underline{64.90} & 54.52 & 75.66 & \textbf{79.30} & \textbf{70.15 (95.35\%)} & 69.82 (82.35\%) \\
            \rowcolor{cyan!15}
            \med~(128) & 19.85 & \underline{84.55} & \underline{60.49} & \underline{87.55} & 65.11 & \underline{78.12} & 45.00 & \textbf{68.15} & \textbf{65.00} & \underline{58.22} & \underline{75.92} & 78.31 & \underline{69.24 (94.11\%)} & \underline{70.81 (83.52\%)} \\
            \midrule
            \multicolumn{15}{c}{\textbf{LoRA Tuning (S2)}}\\
            \midrule
            Dense & 9.58 & 87.84 & 68.09 & 88.99 & 74.59 & 82.21 & 48.60 & 80.32 & 78.69 & 78.18 & 84.65 & 88.77 & 76.16 (103.52\%) & 83.86 (98.91\%) \\
            \midrule
            \rowcolor{gray!15}
            Shortened-ppl & 36.18 & 56.27 & 28.41 & 61.68 & 50.43 & 65.99 & 33.20 & 39.33 & 24.04 & 10.16 & 4.76 & 2.12 & 44.91 (61.05\%) & 5.68 (6.69\%) \\
            \rowcolor{gray!15}
            Shortened-taylor & 1137.25 & 50.92 & 33.95 & 66.17 & 55.48 & 63.60 & 29.80 & 43.17 & 29.91 & 23.29 & 18.25 & 18.57 & 46.62 (63.37\%) & 20.03 (23.63\%) \\
            \rowcolor{gray!15}
            CoopPruner & 27.52 & 61.36 & 33.36 & 61.22 & 50.03 & 69.04 & 33.00 & 43.79 & 24.88 & 14.41 & 13.49 & 6.44 & 47.08 (63.99\%) & 11.44 (13.50\%) \\
            \rowcolor{orange!15}
            DDP & 36.47 & 62.66 & 39.84 & 81.65 & 57.61 & 68.00 & 34.40 & 52.39 & 49.44 & 43.62 & 53.70 & 67.32 & 55.74 (75.77\%) & 54.88 (64.72\%) \\
            \rowcolor{cyan!15}
            SkipGPT & 21.38 & 70.15 & 48.72 & 75.25 & 49.80 & 68.93 & 32.60 & 54.47 & 57.69 & 49.16 & 74.86 & 80.36 & 57.20 (77.74\%) & 68.12 (80.34\%) \\
            \rowcolor{cyan!15}
            \med~(32) & 15.65 & 85.31 & \underline{63.22} & 87.67 & \textbf{68.50} & \underline{78.56} & 47.59 & \underline{71.77} & 67.51 & 57.67 & 78.57 & 83.09 & 71.26 (96.86\%) & 73.11 (86.22\%) \\
            \rowcolor{cyan!15}
            \med~(64) & \underline{15.20} & \underline{85.35} & 62.88 & \textbf{88.31} & 66.69 & \textbf{79.48} & \underline{48.40} & 71.35 & \underline{68.28} & \textbf{60.44} & \textbf{80.68} & \underline{84.68} & \underline{71.34 (96.96\%)} & \textbf{75.26 (88.76\%)} \\
            \rowcolor{cyan!15}
            \med~(128) & \textbf{15.11} & \textbf{86.36} & \textbf{64.07} & \underline{88.25} & \underline{67.56} & 78.34 & \textbf{48.80} & \textbf{72.56} & \textbf{68.66} & \underline{60.25} & \underline{79.10} & \textbf{85.21} & \textbf{71.82 (97.62\%)} & \underline{74.85 (88.28\%)} \\
            \bottomrule
        \end{tabular}
    }
    \label{tbl:full_bench_qwen3-14b_50}
\end{table}

\end{document}